\documentclass{article}

\usepackage[main,final]{neurips_2026}
\usepackage{graphicx}
\usepackage{marvosym} 
\usepackage{float}
\usepackage{amsmath}
\usepackage{multirow}
\usepackage{enumitem}
\usepackage{subcaption}
\usepackage[utf8]{inputenc}
\usepackage{booktabs} 
\usepackage{makecell} 
\usepackage{tabularx} 

\usepackage[utf8]{inputenc} 
\usepackage[T1]{fontenc}    
\usepackage{hyperref}       
\usepackage{url}            
\usepackage{booktabs}       
\usepackage{amsfonts}       
\usepackage{nicefrac}       
\usepackage{microtype}      
\usepackage{xcolor}         
\usepackage{colortbl}       
\hypersetup{hidelinks}

\definecolor{repoRow}{HTML}{F3F3F3}
\definecolor{repoAvg}{HTML}{F7F7F7}
\definecolor{repoGain}{HTML}{0B6B3A}
\definecolor{repoLoss}{HTML}{8A1F1F}
\newcommand{\best}[1]{\textbf{#1}}
\newcommand{\second}[1]{\underline{#1}}
\newcommand{\gain}[1]{\textcolor{repoGain}{\scriptsize (#1)}}
\newcommand{\loss}[1]{\textcolor{repoLoss}{\scriptsize (#1)}}
\newcommand{\na}{\textemdash}
\newcommand{\piZeroTab}{\makecell[l]{$\pi_{0}$ (RSS'25)\\[-1pt]{\tiny \citep{black2024pi_0}}}}
\newcommand{\piHalfTab}{\makecell[l]{$\pi_{0.5}$ (CoRL'25)\\[-1pt]{\tiny \citep{intelligence2025pi_}}}}
\newcommand{\RDTTab}{\makecell[l]{RDT (ICLR'25)\\[-1pt]{\tiny \citep{liu2024rdt}}}}
\newcommand{\GOOneTab}{\makecell[l]{GO-1 (arXiv'25)\\[-1pt]{\tiny \citep{bu2025agibot}}}}

\title{RePO-VLA: Recovery-Driven Policy Optimization for Vision-Language-Action Model}

\author{%
  Weijia Liufu\textsuperscript{1,*}\quad
  Xiaoyu Guo\textsuperscript{2,3,*,\textdagger}\quad
  Ruiyi Chen\textsuperscript{1}\quad
  Jingzhi Liu\textsuperscript{1}\quad
  Kaidong Zhang\textsuperscript{1}\\
  \textbf{Xiwen Liang\textsuperscript{1}\quad
  Jianqi Lin\textsuperscript{1}\quad
  Dawei Sun\textsuperscript{1}\quad
  Yuze Wang\textsuperscript{4}\quad
  Rongtao Xu\textsuperscript{5}}\\
  \textbf{Bingqian Lin\textsuperscript{1}\quad
  Bowen Yang\textsuperscript{6}\quad
  Tongtong Cao\textsuperscript{6}\quad
  Bowen Peng\textsuperscript{3}\quad
  Dongyu Zhang\textsuperscript{1}}\\
  \textbf{Guangrun Wang\textsuperscript{1}\quad
  Min Wang\textsuperscript{2}\quad
  Liang Lin\textsuperscript{1,3,\Letter}\quad
  Xiaodan Liang\textsuperscript{1,3,\Letter}}\\
  \normalfont\textsuperscript{1}Sun Yat-sen University\quad \textsuperscript{2}South China University of Technology\\
  \normalfont\textsuperscript{3}Pengcheng Laboratory\quad \textsuperscript{4}Harbin Institute of Technology, Shenzhen\\
  \normalfont\textsuperscript{5}Institute of Automation, Chinese Academy of Sciences\quad \textsuperscript{6}Huawei Noah's Ark Lab\\
  \normalfont\textsuperscript{*}Equal contribution.\quad
  \textsuperscript{\textdagger}Project leader.\quad
  \textsuperscript{\Letter}Corresponding authors.
}

\begin{document}

\maketitle
\begingroup
  \renewcommand{\thefootnote}{\Letter}
  \footnotetext[3]{Correspondence:
    \href{mailto:linlng@mail.sysu.edu.cn}{\texttt{linlng@mail.sysu.edu.cn}},
    \href{mailto:liangxd9@mail.sysu.edu.cn}{\texttt{liangxd9@mail.sysu.edu.cn}}.}
\endgroup

\begin{abstract}
Vision-Language-Action (VLA) models remain brittle in long-horizon, contact-rich manipulation because success-only imitation provides little supervision for execution drift, while failed rollouts are often discarded. We introduce RePO-VLA, a recovery-driven policy optimization framework that assigns distinct roles to success, recovery, and failure trajectories. RePO-VLA first applies Recovery-Aware Initialization (RAI), slicing each corrective suffix into a new episode so corrective actions depend on the current adverse state rather than the preceding failure. It then learns a Progress-Aware Semantic Value Function (PAS-VF), aligning spatiotemporal trajectory features with instructions and successful references. The resulting labels salvage useful failure prefixes via reliability decay, while low-value labels mark drift and terminal breakdowns, teaching differences among nominal, failed, and corrective actions. The data engine turns adverse states into planner-generated or human-collected corrective rollouts, teaching recovery to the success manifold. Value-Conditioned Refinement (VCR) trains the policy to prefer high-progress actions. At deployment, a fixed high-value token biases actions toward the learned success manifold without online failure detectors or heuristic retries. We introduce the Failure-Recovery Benchmark (FRBench), with standardized error injection and recovery-focused evaluation. Across simulated and real-world bimanual tasks, RePO-VLA improves robustness, raising adversarial success from 20\% to 75\% on average and up to 80\% in scaled real-world trials. Code and video results are available at \url{https://1018weijia.github.io/RePO-VLA-Web/}.
\end{abstract}

\section{Introduction}
Vision-Language-Action (VLA) models have advanced general-purpose robot control \citep{kim2024openvla, black2024pi_0}, yet a persistent gap remains between nominal benchmark success and reliable long-horizon execution. The gap is especially pronounced in contact-rich bimanual manipulation, where small deviations in grasp pose, object contact, or inter-arm timing can quickly derail a task. These perturbations rarely make the task impossible; instead, they create recoverable adverse states requiring targeted correction before nominal progress can continue. Resilient execution therefore has a temporal structure: useful approach behavior, drift or stalled progress, and correction back to the success manifold.

\begin{figure}[t]
\centering
\includegraphics[width=\linewidth]{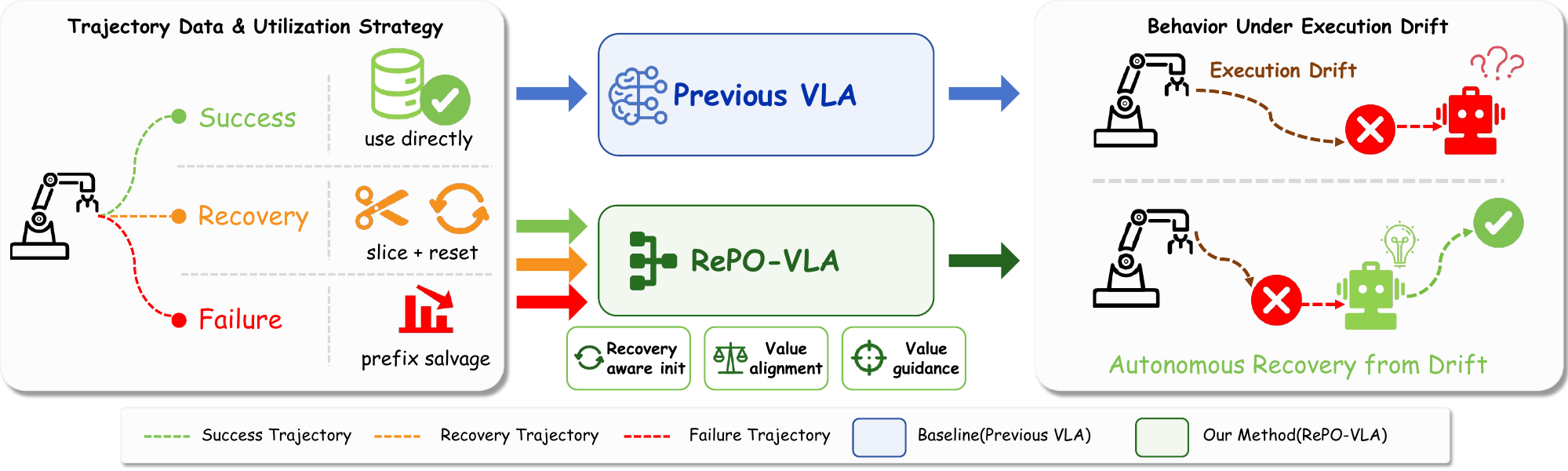}
\caption{\textbf{Recovery-driven trajectory utilization.} Prior VLAs mainly imitate successful demonstrations and often fail under execution drift. RePO-VLA assigns distinct supervision to success, recovery, and failure trajectories, enabling autonomous correction from adverse states.}
\label{fig:Intro}
\end{figure}

Current supervision provides little signal for this behavior, despite growing attention to robotic failures \citep{liu2023reflect, lu2025robofac, haddadin2017robot}. Human-in-the-loop correction \citep{shi2024yell, wu2025robocopilot, pramanick2022talk} and VLM/LLM replanning \citep{huang2022inner, shinn2023reflexion, guo2024doremi} can recover from some high-level mistakes, but they are costly and often disconnected from low-level contact dynamics \citep{hafner2023mastering, ahn2022can}. In practice, most VLA training still relies on supervised fine-tuning (SFT) over successful demonstrations. This creates a failure-utilization blind spot: pure failures are discarded even though their early portions often contain valid approach, contact preparation, or partial-manipulation priors. At the same time, their late portions indicate where behavior drifts away from the success manifold and should be marked as low-value rather than imitated. Recovery demonstrations pose a complementary ambiguity because they contain failed history, adverse states, and corrective actions. Naively imitating the whole sequence conflates the distribution that causes drift with the distribution that repairs it, leading to causal confusion and mode averaging. Sparse binary rewards compound the issue by failing to distinguish useful approach, temporary stagnation, active recovery, and terminal failure.

Figure~\ref{fig:Intro} summarizes the core mismatch: success-only training leaves prior VLAs brittle under drift, while robust recovery requires assigning different roles to nominal-success, recovery, and failure trajectories. Failed rollouts should be decomposed rather than discarded, especially as VLA pipelines converge around large-scale imitation and flow-matching backbones. Portions close to nominal execution can be preserved, later portions show where the same trajectory becomes disadvantageous, and adverse states from injected or policy-induced failures can be turned into generated or collected corrective rollouts.

We propose RePO-VLA, a recovery-driven policy optimization framework that learns from nominal-success, failed, and recovered rollouts under a unified progress signal. It is a recovery-data and supervision framework used at fine-tuning time. It does not modify the pretrained VLA backbone, and it does not claim that recovery representations emerge during large-scale pretraining. A backbone that models failure, recovery, and progress directly is left as future work.

RePO-VLA first performs \textbf{Recovery-Aware Initialization (RAI)}. A raw recovery episode mixes the failure prefix that causes an adverse state with the later corrective suffix. Imitating that whole episode can teach the prefix as a trigger for correction. That causal confusion makes the original mistake look necessary, so later errors compound instead of being repaired from the current state. RAI slices out the corrective suffix and serializes it as a new episode, so the corrective skill is conditioned on the adverse state itself. It then learns a \textbf{Progress-Aware Semantic Value Function (PAS-VF)} by aligning spatiotemporal video features with language instructions. PAS-VF assigns dense labels to mixed-quality data, isolates error segments from correction segments, and preserves useful early portions of failed trajectories through reliability decay. These labels teach distributional differences among nominal progress, failed drift, and corrective recovery, enabling value-differentiated actions instead of one averaged imitation mode. Finally, \textbf{Value-Conditioned Refinement (VCR)} trains the policy to follow high-value behavior. During inference, RePO-VLA conditions on a fixed high-value token and uses the learned success manifold as an attractor, without online failure detectors or hand-coded retry rules.

To evaluate recovery as a first-class capability, we introduce the Failure-Recovery Benchmark (FRBench), with standardized error injection, recovery-focused protocols, and simulated and real-world bimanual tasks. FRBench separates nominal execution from recovery execution, making it possible to measure whether a policy can actively repair execution drift rather than merely succeed under clean conditions. Its evaluation stages---Nominal, Error, and Recovery---condition recovery on a verified adverse state, preventing improvements in reaching or grasping from being mistaken for true correction. These evaluation stages are separate from the RAI and VCR training stages. Across FRBench-Sim and real-world adversarial perturbations, RePO-VLA improves recovery success over imitation baselines and shows an empirical recovery-data scaling trend as data density increases.

Our contributions are:
\begin{itemize}[leftmargin=*, topsep=0pt, itemsep=0pt, parsep=0pt, partopsep=0pt]
    \item \textbf{RePO-VLA}, an RAI and VCR supervision framework rather than a new VLA backbone. It eliminates recovery-data causal confusion and structurally reuses failed rollouts: progress alignment preserves useful failure prefixes, low-value labels expose drift-to-correction boundaries, and the data engine converts adverse failure states into generated or collected corrective rollouts for value-conditioned recovery.
    \item \textbf{FRBench}, a recovery benchmark with structured error injection and progress-sensitive evaluation in simulation and real-world bimanual settings.
    \item \textbf{Empirical validation} showing strong recovery gains over SFT baselines and improved recovery with increased recovery-data density.
\end{itemize}

\section{Related Work}

\subsection{Vision-Language-Action Models}
VLA models connect visual-semantic reasoning with robot control. OpenVLA \citep{kim2024openvla} casts actions as language tokens, while recent $\pi$-series models \citep{black2024pi_0, intelligence2025pi_} use continuous generative policies for dexterous trajectories. Long-horizon extensions add multimodal reasoning \citep{wen2025dvla}, memory \citep{shi2025memoryvla}, or RL fine-tuning \citep{guo2025improving, chen2025pi_, intelligence2025pi, zhu2025wmpo}. However, most pipelines remain success-demonstration centric, leaving policies poorly supervised for recovery from real perturbations.

\subsection{Failure Recovery on Robotic Manipulation}
Recovery methods include modular replanning, interactive correction, and learned refinement. VLM-guided systems such as ReplanVLM \citep{mei2024replanvlm}, REPLAN \citep{skreta2024replan}, RACER \citep{dai2025racer}, and RoboFAC \citep{lu2025robofac} detect precondition errors \citep{raman2024cape} or generate recovery instructions, while interactive methods such as ThriftyDAgger \citep{hoque2021thriftydagger} and TRANSIC \citep{jiang2024transic} reduce distribution shift through intervention; both remain largely external to the learned low-level policy.

Learning-based recovery internalizes this capability. Failure and vulnerability predictors such as FIPER \citep{romer2025failure} and RoboMD \citep{sagar2024robomd} estimate risk, while RLFT methods including RLPD \citep{ball2023efficient}, GRAPE \citep{zhang2024grape}, and ConRFT \citep{chen2025conrft} align policies using outcome feedback. These signals are usually sparse or binary \citep{nakamoto2023cal, mandlekar2020human}. FailSafe \citep{lin2025failsafe} synthesizes recovery data but still uses SFT, which does not separate erroneous states from corrective actions. RePO-VLA instead uses dense progress labels to train value-conditioned recovery.

Our work is closest to methods that use imperfect or corrective data, but differs by structuring supervision rather than treating each episode uniformly: RePO-VLA assigns low value to error regions, high value to corrective regions, and reliability-decayed progress to useful failure prefixes. This separation makes recovery learning compatible with mixed-quality datasets where success, failure, and repair behaviors coexist; extended discussion is provided in \hyperref[sec:appendix_more_related_work]{Appendix~\ref*{sec:appendix_more_related_work}}.

Prior robustness methods share the goal but not the supervision. FailSafe, the closest, synthesizes failure-recovery data for VLA training; RACER and RoboFAC focus on failure reasoning or replanning; ThriftyDAgger on intervention timing; TRANSIC on a gated residual correction policy; and RLFT-style methods and $\pi^{*}_{0.6}$-RECAP on broader improvement from rewards or mixed-quality experience. RePO-VLA instead separates failure prefixes, error regions, and corrective suffixes within one low-level policy. We claim no superiority over methods without head-to-head comparison.

\section{RePO-VLA}
\begin{figure*}[htbp]
\centering
\includegraphics[width=\linewidth]{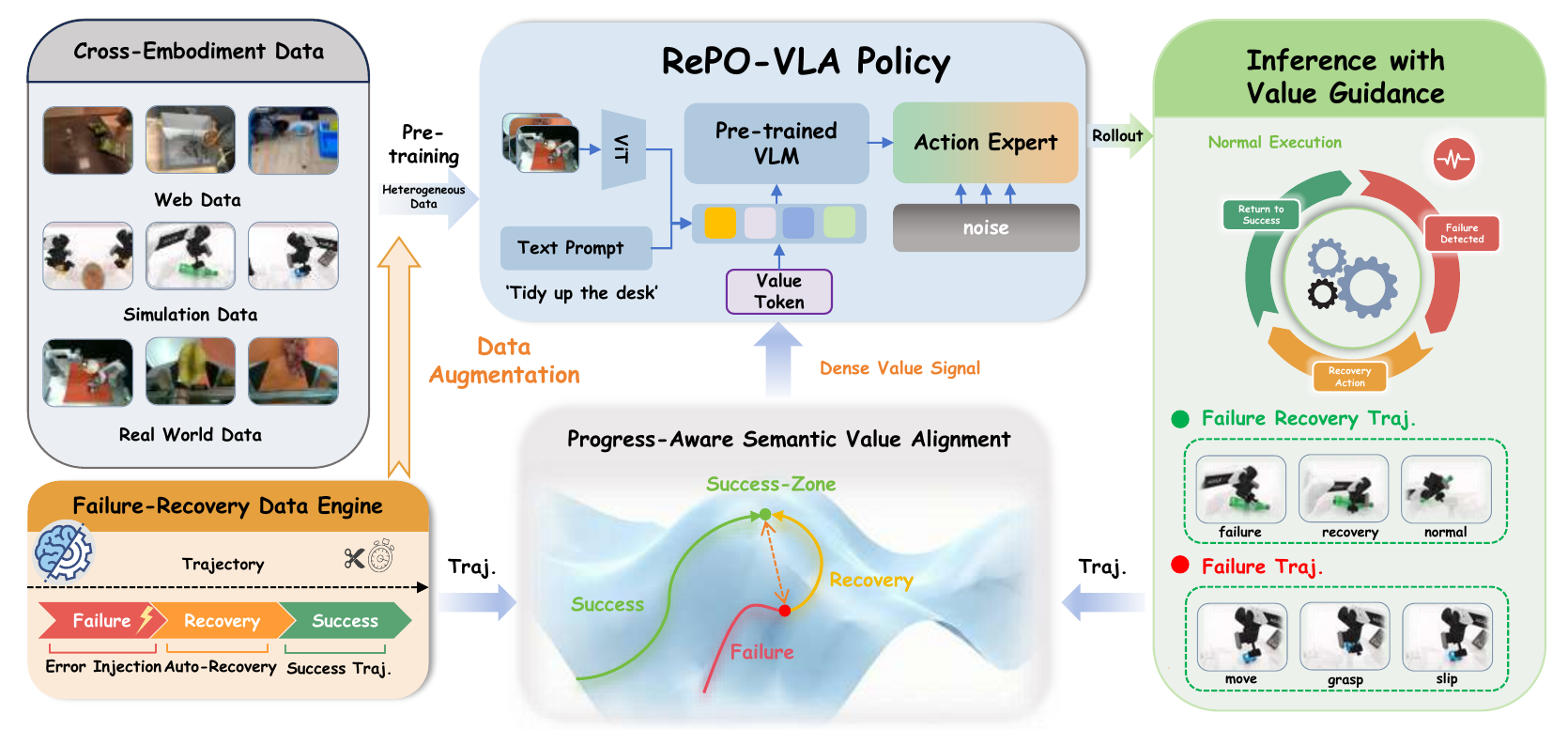}
\caption{\textbf{Overview of RePO-VLA.} The framework builds recovery data, learns a progress-aware semantic value signal, and refines a value-conditioned VLA policy. At inference, constant high-value conditioning ($v=1.0$) steers execution back toward the success manifold after drift.}
\label{fig:Framework}
\end{figure*}

RePO-VLA uses two training stages and does not modify the pretrained backbone. \textbf{Recovery-Aware Initialization (RAI)} fine-tunes on expert demonstrations and corrective suffixes serialized as new episodes, so a failure prefix is not part of the same episode as the correction. \textbf{Value-Conditioned Refinement (VCR)} then trains on the original mixed trajectories with dense progress labels for nominal-success, failure, and recovery behavior. The same policy handles nominal execution and recovery without switching modules at test time. In the experiments, Phase I names the RAI checkpoint and Full names the VCR checkpoint. Those names are training stages, distinct from the Nominal, Error, and Recovery evaluation stages.

\subsection{Recovery-Aware Initialization (RAI)}
RAI initializes the policy with expert demonstrations and recovery actions while avoiding causal confusion \citep{NEURIPS2019_94701864}. A raw recovery trajectory contains two qualitatively different regions: the failure prefix that leads the system away from progress, and the corrective segment that brings it back. The base policy consumes a single-timestamp observation and does not take stacked historical image frames. If the failure prefix and the correction are still stored as one episode, the policy can treat the preceding mistake as the context that triggers correction; see \hyperref[sec:appendix_history_reset]{Appendix~\ref*{sec:appendix_history_reset}} for details.

We address this with Trajectory Slicing with Episode Reset (TSER). Given a recovery that starts at $t_{rec}$, we discard the failure prefix and extract $\tau'=\{(o_t,a_t)\}_{t=t_{rec}}^T$ as an independent correction episode whose time index restarts at $t=0$. No historical image frames are attached before this onset. The resulting dataset is denoted $\mathcal{D}_{rec}^{reset}$. These samples expose the policy to adverse states such as an empty gripper, shifted object, or misaligned bimanual pose, but remove the failure prefix from the same episode. Thus RAI learns recovery as a state-conditioned skill rather than a replay of a particular failed rollout. TSER is dataset-level episode bookkeeping, not a reset of an observation-history buffer. Deployment uses the same single-timestamp observation, as clarified in \hyperref[sec:appendix_history_treatment]{Appendix~\ref*{sec:appendix_history_treatment}}.

RAI mixes expert data $\mathcal{D}_{expert}$ and reset recovery data $\mathcal{D}_{rec}^{reset}$:

\begin{equation}
\mathcal{L}_{SFT}=-\mathbb{E}_{\tau\sim\mathcal{D}_{expert}}\sum_t\log\pi(a_t|o_t)-\lambda\,\mathbb{E}_{\tau'\sim\mathcal{D}_{rec}^{reset}}\sum_t\log\pi(a_t|o_t),
\end{equation}
where each sum follows the valid horizon of its trajectory, and $\lambda$ balances nominal imitation and recovery learning.

\subsection{Progress-Encoded Semantic Value Learning (PAS-VF)} \label{sec:value_function}
VCR requires dense labels that preserve useful pre-failure behavior without rewarding terminal breakdowns. A simple zero label for a failed trajectory is too coarse: many failures contain a correct approach, stable pre-contact motion, or partial manipulation before the final deviation. Conversely, assigning a high label to the whole trajectory would reward the terminal error. We therefore learn a self-referential Progress-Aware Semantic Value Function (PAS-VF), denoted $V(\tau)$, that estimates task progress as proximity to successful trajectories in a semantic latent space.

\textbf{Spatiotemporal Alignment.} We map trajectories and language instructions into a shared manifold $\mathcal{Z}$. This is important because bimanual tasks can be visually redundant: different arm assignments or grasp poses may represent the same semantic progress. A frozen V-JEPA spatiotemporal encoder $E_v$ \citep{assran2025v} extracts trajectory dynamics and a frozen CLIP text encoder $E_t$ \citep{radford2021learning} encodes the instruction, while lightweight adapters produce $z^v=f_\theta(E_v(\tau))$ and $z^l=g_\phi(E_t(l))$. The base encoders remain frozen; only the adapters are optimized, keeping the value model lightweight and focused on progress alignment rather than representation relearning.

\begin{figure}[htbp]
\centering
\includegraphics[width=\linewidth]{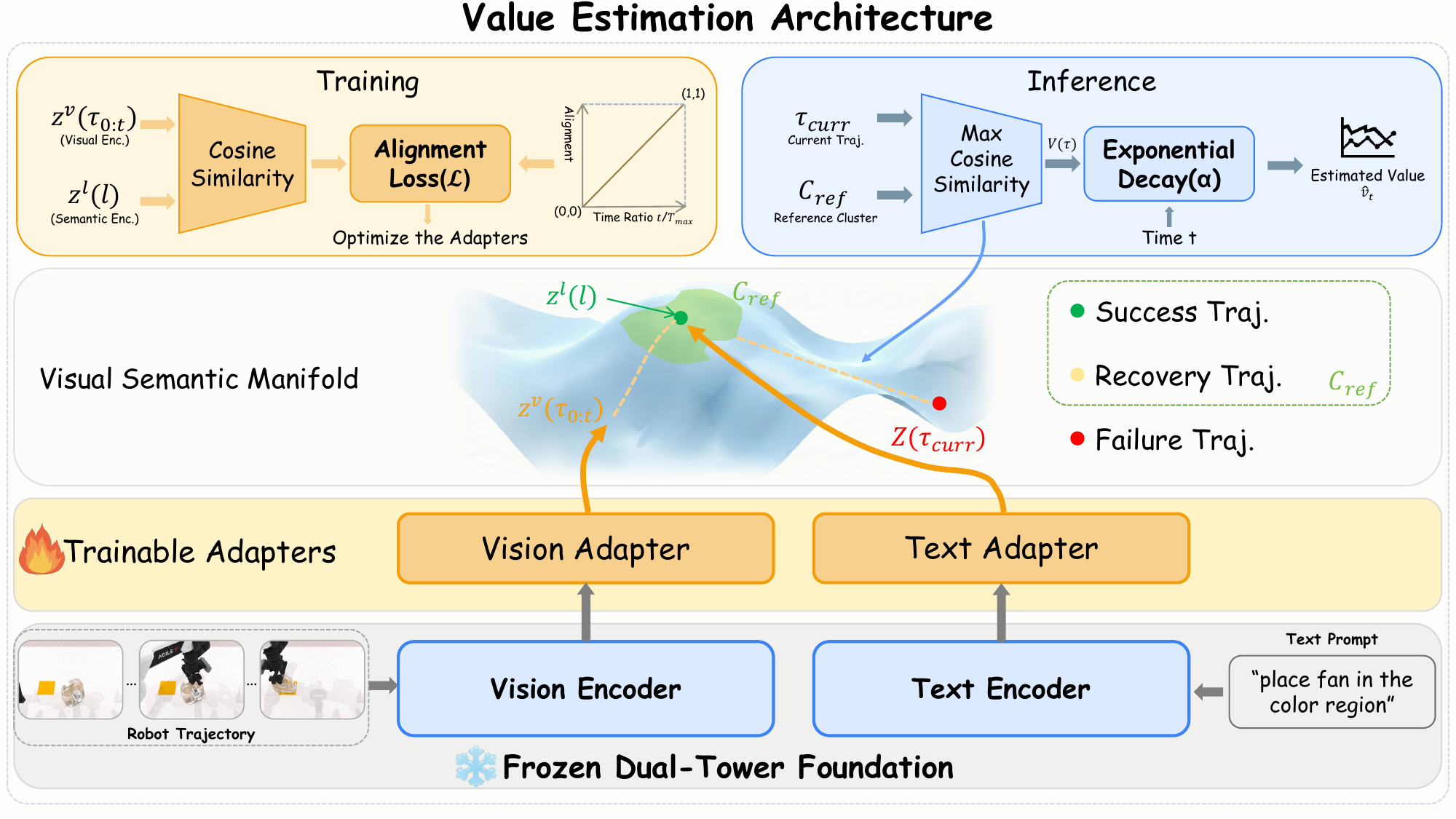}
\caption{\textbf{Semantic value alignment.} Frozen visual and language encoders are projected into $\mathcal{Z}$; cosine similarity provides a dense progress signal.}
\label{fig:dual_tower}
\end{figure}

\textbf{Training: Monotonic Progress Alignment.} 
Adapters are trained on successful trajectories so cosine similarity tracks normalized temporal progress. For each prefix $\tau_{0:t}$, the target progress is $t/T_\tau$, which encourages early prefixes to remain far from the instruction embedding and completed prefixes to align closely with it. With $z_t^v=f_\theta(E_v(\tau_{0:t}))$:
\begin{equation}
\mathcal{L}_{align} = \mathbb{E}_{\tau \in \mathcal{T}_{succ}} \left[ \frac{1}{T_\tau} \sum_{t=1}^{T_\tau} \left( \text{CosSim}(z_t^v, z^l) - \frac{t}{T_\tau} \right)^2 \right].
\end{equation}

\textbf{Self-Referential Progress Estimation.} After alignment, we use successful executions as an internal reference set. For an unlabeled failure, progress is the nearest similarity to a reference cluster $\mathcal{C}_{ref}$ of successful embeddings:

\begin{equation}
V(\tau_{fail}) = \max_{z \in \mathcal{C}_{ref}} \text{CosSim}(z_{\tau_{fail}}^v, z).
\end{equation}

\begin{figure*}[t]
\centering
\begin{minipage}{0.48\textwidth}\centering\includegraphics[width=\textwidth]{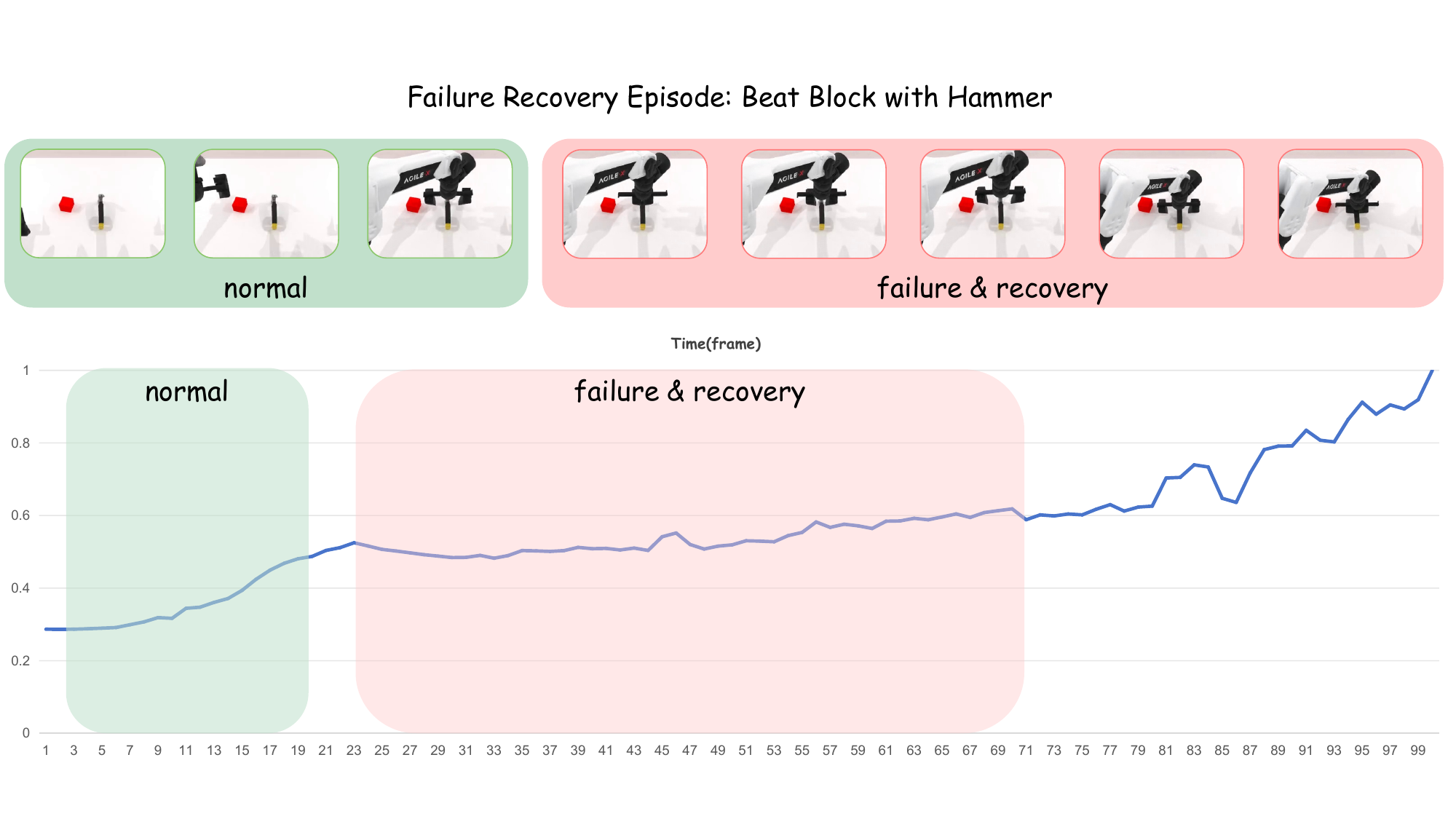}
\subcaption{Failure Recovery Episode value modeling}
\label{fig:value_function_1}
\end{minipage}
\hfill
\begin{minipage}{0.48\textwidth}\centering\includegraphics[width=\textwidth]{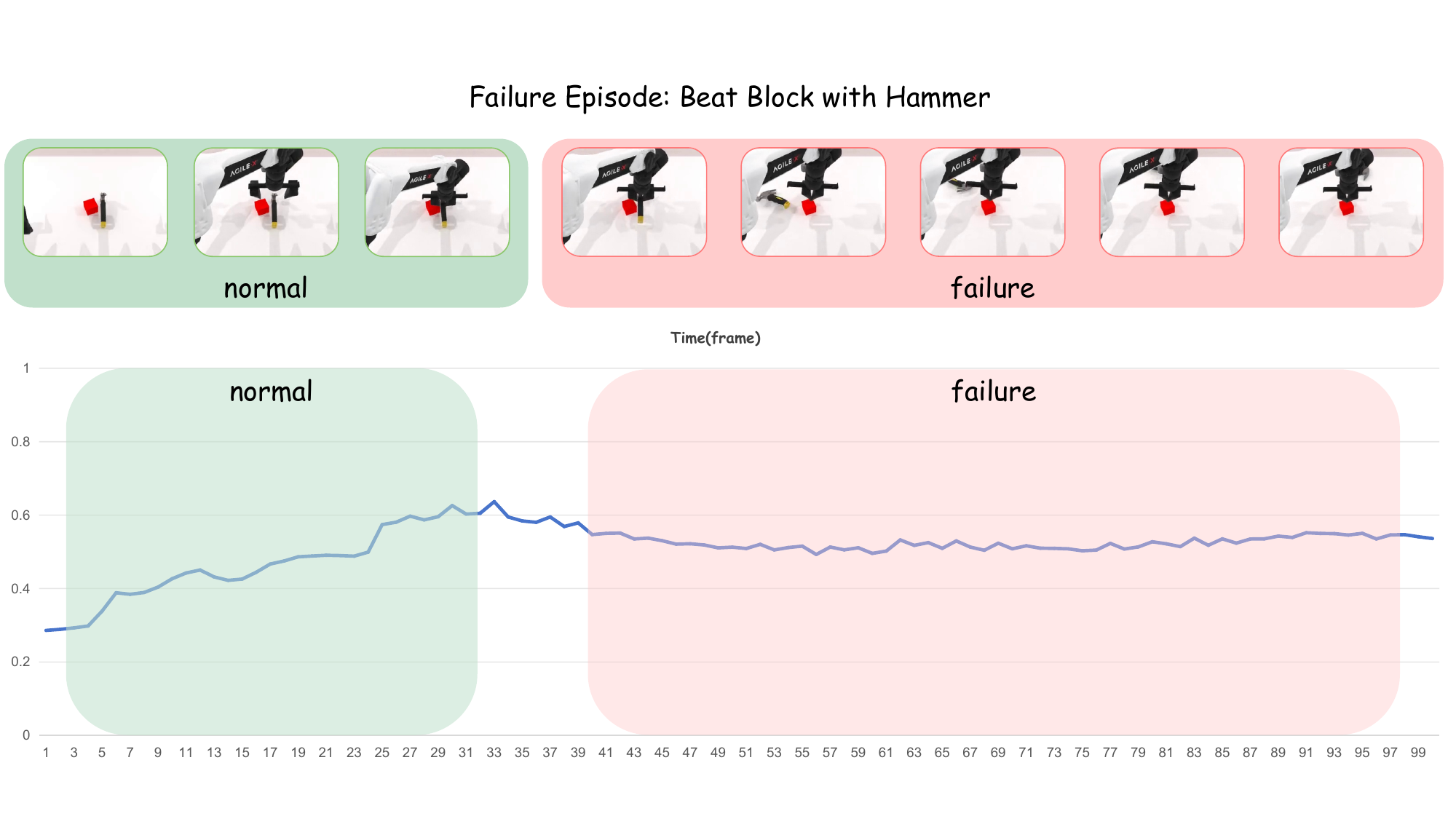}
\subcaption{Failure Episode value modeling}
\label{fig:value_function_2}
\end{minipage}
\caption{\textbf{Visualization of the Progress-Aware Value Landscape.}}
\label{fig:value_function}
\end{figure*}

Here $z_{\tau_{fail}}^v=f_\theta(E_v(\tau_{fail}))$. This self-referential score avoids manual progress annotation and remains task-relative: a trajectory is valuable when it approaches the successful manifold for its own instruction. Figure~\ref{fig:value_function} visualizes the resulting progress, plateau, and failure regimes.


\subsection{Value-Conditioned Refinement (VCR)}
VCR uses $V(\tau)$ to separate high-value correction from low-value failure behavior across the mixed dataset. Unlike RAI, which trains on sliced corrective suffixes in $\mathcal{D}_{rec}^{reset}$, VCR trains on the original mixed trajectories $\mathcal{D}_{rec}^{raw}$. Failure-inducing and corrective actions therefore remain in one episode and are distinguished by value labels rather than by an episode boundary. The observation at each step is still a single timestamp; additional episode-boundary details are in \hyperref[sec:appendix_history_treatment]{Appendix~\ref*{sec:appendix_history_treatment}}.

\subsubsection{Progress-Aware Hindsight Labeling}

We construct a continuous value landscape over the raw mixed dataset by assigning each frame a dense label $v_t \in [0,1]$. The labels are designed to separate behavior quality rather than trajectory identity:
\begin{itemize}[leftmargin=*, topsep=0pt, itemsep=0pt, parsep=0pt, partopsep=0pt]
\item \textbf{Success and recovery:} successful trajectories $\tau_{succ}$ and effective recovery suffixes $\tau_{rec}^{+}\subset\mathcal{D}_{rec}^{raw}$ receive $v_t=1.0$, since these frames contain actions that either maintain or restore progress.

\item \textbf{Error segments:} deviating recovery prefixes $\tau_{rec}^{-}\subset\mathcal{D}_{rec}^{raw}$ receive $v_t=0.0$. This explicitly tells the policy that the actions leading into the adverse state should not be averaged with the corrective actions that follow, even though both occur in the same raw episode.

\item \textbf{Pure failures:} failed rollouts $\tau_{fail}$ keep early kinematic priors through reliability decay,
\begin{equation}
v_t = V(\tau_{fail}) \cdot \left( 1 - \frac{t}{T} \right)^\alpha, \quad \forall t \in \tau_{fail},
\end{equation}
with $\alpha=3.0$. This retains useful early motion while rapidly down-weighting states near irreversible failure; see \hyperref[sec:appendix_decay]{Appendix~\ref*{sec:appendix_decay}}.


\end{itemize}

\subsubsection{Goal-Conditioned Architecture and Inference}

\textbf{Value-Conditioned Training.} We augment the flow-matching $\pi_{0.5}$ policy \citep{intelligence2025pi_} with a value token $\mathbf{e}_{val}=\text{MLP}_{val}(v_t)$. The transformer attends to this token together with visual and language tokens, allowing the same observation to map to different action modes depending on desired progress. In this paper the normalized success anchor is the high-value token $v=1.0$; that number is the top of our progress labels, not a task-independent physical scale. We fine-tune on $\mathcal{D}_{total}=\mathcal{D}_{succ}\cup\mathcal{D}_{rec}^{raw}\cup\mathcal{D}_{fail}$:
\begin{equation}
    \mathcal{L}_{VCR} = - \mathbb{E}_{\tau \sim \mathcal{D}_{total}} \left[ \sum_{t=1}^{T_\tau} \log \pi_\theta(a_t | o_t, l, \mathbf{e}_{val}(v_t)) \right].
\end{equation}

\textbf{Goal-Conditioned Autonomous Recovery.} During deployment, we do not run an online failure detector, estimate $V(\tau)$, or switch policies. The policy receives the current observation and a fixed high-value token, set to the normalized success anchor $v=1.0$, at every step. Nominal and corrective actions come from this same interface. \hyperref[sec:appendix_no_detector]{Appendix~\ref*{sec:appendix_no_detector}} further discusses the detector-free inference setting.

\section{FRBench}

\begin{figure*}[t]
    \centering
    \includegraphics[width=\linewidth]{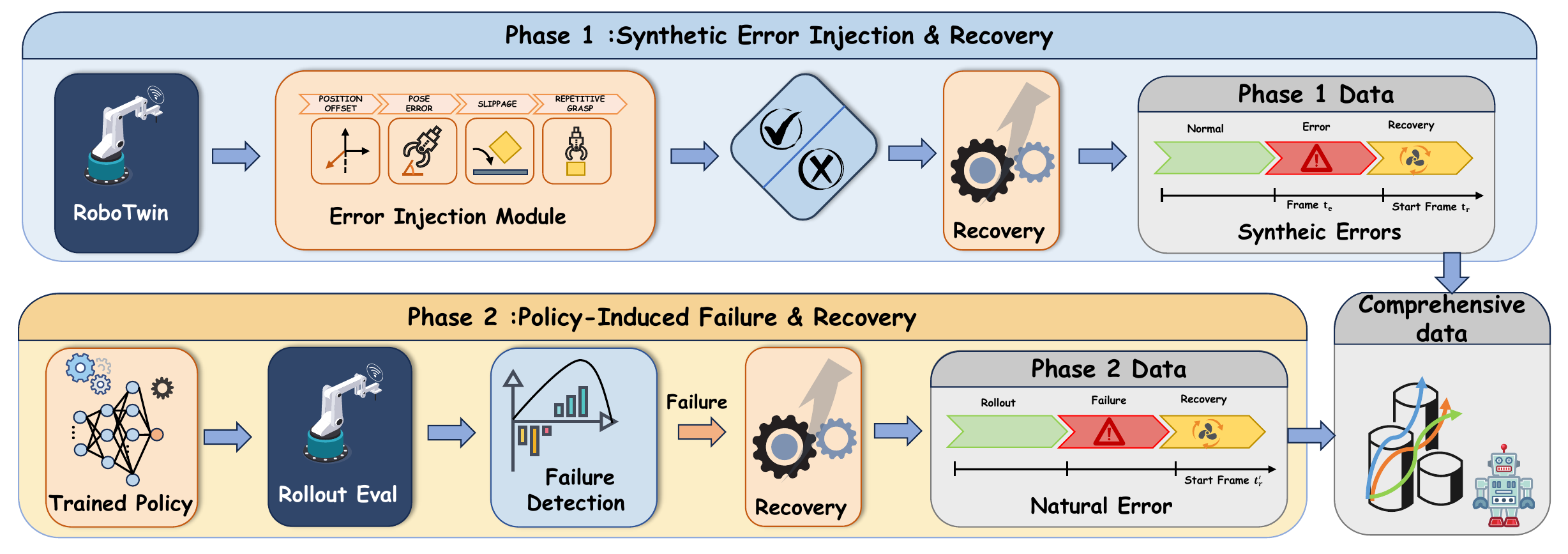}
    \caption{\textbf{Failure-Recovery Data Engine.} Recovery data combines control-intercepted error injection with policy-induced rollouts.}
    \label{fig:data_engine}
\end{figure*}

FRBench evaluates whether bimanual policies can restore task progress after physical breakdowns. Unlike success-only suites, it separates nominal proficiency from corrective capability through controlled adverse-state projection. FRBench spans RoboTwin simulation \citep{chen2025robotwin} and real-world tasks, with 23,453 simulated episodes over 46 tasks; detailed protocol and dataset statistics are provided in \hyperref[sec:appendix_frbench_protocol]{Appendix~\ref*{sec:appendix_frbench_protocol}} and \hyperref[sec:appendix_frbench_stats]{Appendix~\ref*{sec:appendix_frbench_stats}}.

\noindent\textbf{Phase-based protocol.} FRBench evaluates recovery as a controlled state-transition problem: each trial passes through nominal execution, error projection, and recovery execution, so recovery success is conditioned on a verified adverse state rather than confounded by approach quality.

\begin{table}[t]
\centering
\caption{\textbf{FRBench phase-based evaluation protocol.} The protocol isolates nominal task competence from corrective behavior after a controlled physical breakdown.}
\label{tab:frbench_protocol}
\scriptsize
\setlength{\tabcolsep}{3pt}
\renewcommand{\arraystretch}{1.05}
\begin{tabularx}{\linewidth}{@{}l l X X@{}}
\toprule
Phase & State & Control or intervention & Isolated measurement \\
\midrule
Nominal & $\phi_t=\textsc{Nominal}$ & Policy executes under clean or randomized conditions. & Baseline reaching, grasping, and task progress. \\
Error & $\phi_t=\textsc{Error}$ & Structured E1--E4 perturbation projects the scene into an adverse state. & Reproducible physical breakdown independent of policy scoring. \\
Recovery & $\phi_t=\textsc{Recovery}$ & The policy continues from the adverse state without external retry logic. & Corrective actions that restore task progress. \\
\bottomrule
\end{tabularx}
\vspace{-1mm}
\end{table}

\subsection{Failure-Recovery Data Engine}
The data engine in Fig.~\ref{fig:data_engine} supplies recovery supervision through two complementary sources, balancing broad coverage with policy-specific realism:

\noindent \textbf{Interception-Based Synthetic Injection.} We hook expert-planner execution nodes and inject targeted errors (E1--E4) during critical segments such as grasping or lifting. A hierarchical planner then restores the task state. The injection sequence and perturbation operators are detailed in \hyperref[sec:appendix_frbench_sequence]{Appendix~\ref*{sec:appendix_frbench_sequence}} and \hyperref[sec:appendix_frbench_taxonomy]{Appendix~\ref*{sec:appendix_frbench_taxonomy}}.

\noindent\textbf{Policy-Induced Rollouts.} We also collect closed-loop failures from trained base policies and let the expert planner intervene, yielding in-distribution recovery data for model-specific drift; detection and real-world recovery details appear in \hyperref[sec:appendix_frbench_detection]{Appendix~\ref*{sec:appendix_frbench_detection}}.

Both sources are merged into raw recovery trajectories $\mathcal{D}_{rec}^{raw}$ for VCR; RAI derives $\mathcal{D}_{rec}^{reset}$ from them via TSER. The synthetic stream provides controlled coverage of common physical breakdowns, while the policy-induced stream captures the errors that actually arise from learned visual perception and action generation.

\noindent\textbf{Simulation versus real collection.} In FRBench-Sim, an interception module injects an error and a privileged hierarchical planner synthesizes the correction, yielding 6,392 recovery episodes without a human. In FRBench-Real, adverse states are induced manually and teleoperated back to success, since friction, contact, fluids, and deformables are hard to script. Neither is used at deployment: one policy with a fixed high-value token emits nominal and corrective actions, with no online value estimate, failure detector, or retry controller.

\section{Experiment}
\subsection{Simulation Evaluation (FRBench-Sim)}
\label{sec:sim_exp}

We evaluate FRBench-Sim on Dynamic Grasp Failure, a proxy for contact anomalies such as \texttt{grasp\_slip} and \texttt{premature\_close}. This perturbation is deliberately low-level: it does not change the task instruction, but it breaks object control and requires the policy to re-enter a valid grasping state. The full perturbation taxonomy is given in \hyperref[sec:appendix_frbench_taxonomy]{Appendix~\ref*{sec:appendix_frbench_taxonomy}}. Baselines include RDT \citep{liu2024rdt}, GO-1 \citep{bu2025agibot}, $\pi_0$ \citep{black2024pi_0}, $\pi_{0.5}$ \citep{intelligence2025pi_}, and the RAI checkpoint, which the tables label Phase I. These imitation baselines are trained only on optimal nominal-success demonstrations. Failed, suboptimal, and recovery trajectories are removed, so a low recovery score is not caused by imitating corrupted actions. It is what success-only imitation does after the state leaves the nominal demonstration distribution.

%
\noindent\textbf{Protocol.} We run 50 rollouts per task in RoboTwin Clean and Random settings. Recovery trials inject a grasp disturbance by holding the gripper open for 30 frames ($\sim$1s) at grasp initiation, forcing the policy to re-establish object control; protocol details and visual examples are in \hyperref[sec:appendix_frbench_protocol]{Appendix~\ref*{sec:appendix_frbench_protocol}} and \hyperref[app:data_viz]{Appendix~\ref*{app:data_viz}}.

\begin{table}[t]
\centering
\caption{\textbf{FRBench-Sim success rate (\%).} We report nominal and injected-failure success over ten RoboTwin tasks. Method labels include venue year and citation. Throughout the result tables, $\uparrow$ means higher is better, gray marks RePO-VLA, bold/underline mark best/second-best results, and green/red values denote gains/losses. Here $\Delta$ reports average gain over the corresponding $\pi_{0.5}$ baseline.}
\label{tab:sim_results}
\scriptsize
\setlength{\tabcolsep}{2.35pt}
\renewcommand{\arraystretch}{0.92}
\resizebox{\linewidth}{!}{
\begin{tabular}{llcccccccccccc}
\toprule
Model & Split & \multicolumn{10}{c}{Task Success Rate (\%) $\uparrow$} & Avg. $\uparrow$ & $\Delta$ Avg. $\uparrow$ \\
\cmidrule(lr){3-12}
 & & \makecell{Blocks\\RGB} & \makecell{Blocks\\Size} & \makecell{Hang\\Mug} & \makecell{Lift\\Pot} & \makecell{Move\\Stap.} & \makecell{Open\\Laptop} & \makecell{Place\\Bread} & \makecell{Place\\Can} & \makecell{Place\\Mouse} & \makecell{Press\\Stap.} & & \\
\midrule
\multicolumn{14}{l}{\textbf{Standard (nominal)}} \\
\multirow{2}{*}{\RDTTab} & Clean & 3 & 0 & 23 & 72 & 2 & 59 & 5 & 19 & 1 & 41 & 22.5 & \na \\
& Rand. & 0 & 0 & 16 & 9 & 0 & 32 & 1 & 6 & 0 & 24 & 8.8 & \na \\
\cmidrule(lr){1-14}
\multirow{2}{*}{\GOOneTab} & Clean & 7 & 2 & 0 & 92 & 3 & 65 & 2 & 29 & 15 & 66 & 28.1 & \na \\
& Rand. & 3 & 2 & 0 & 92 & 4 & 60 & 1 & 37 & 10 & 51 & 26.0 & \na \\
\cmidrule(lr){1-14}
\multirow{2}{*}{\piZeroTab} & Clean & 19 & 7 & 11 & 84 & 0 & 85 & 23 & 41 & 7 & 62 & 33.9 & \na \\
& Rand. & 5 & 1 & 3 & 36 & 2 & 46 & 1 & 5 & 1 & 29 & 12.9 & \na \\
\cmidrule(lr){1-14}
\multirow{2}{*}{\piHalfTab} & Clean & 30 & 20 & 6 & 28 & 10 & 52 & 26 & 30 & 12 & 60 & 27.4 & \na \\
& Rand. & 48 & 22 & 10 & 20 & 16 & 74 & 44 & 20 & 28 & 54 & \second{33.6} & \na \\
\cmidrule(lr){1-14}
\rowcolor{repoRow}
\multirow{2}{*}{\makecell[l]{w/o Fail\\(Phase I)}} & Clean & 82 & 54 & 4 & 38 & 26 & 62 & 34 & 48 & 38 & 60 & \best{44.6} & \gain{+17.2} \\
\rowcolor{repoRow}
& Rand. & 68 & 56 & 8 & 36 & 18 & 54 & 50 & 46 & 48 & 56 & \best{44.0} & \gain{+10.4} \\
\midrule
\multicolumn{14}{l}{\textbf{Injected failure}} \\
\multirow{2}{*}{\piHalfTab} & Clean & 20 & 2 & 0 & 6 & 0 & 64 & 0 & 0 & 0 & 58 & \second{15.0} & \na \\
& Rand. & 22 & 6 & 0 & 0 & 0 & 68 & 0 & 0 & 4 & 54 & \second{15.4} & \na \\
\cmidrule(lr){1-14}
\rowcolor{repoRow}
\multirow{2}{*}{\makecell[l]{RePO-VLA\\Full}} & Clean & 70 & 60 & 0 & 30 & 0 & 90 & 20 & 10 & 10 & 80 & \best{37.0} & \gain{+22.0} \\
\rowcolor{repoRow}
& Rand. & 80 & 40 & 20 & 0 & 30 & 100 & 60 & 10 & 20 & 70 & \best{43.0} & \gain{+27.6} \\
\bottomrule
\end{tabular}}
\vspace{0.2mm}

{\scriptsize \emph{Tasks.} Blocks RGB/Size: Blocks Ranking RGB/Size; Move Stap.: Move Stapler Pad; Place Bread/Can/Mouse: Place Bread Skillet/Place Can Basket/Place Mouse Pad; Press Stap.: Press Stapler.}
\end{table}

Table~\ref{tab:sim_results} shows two effects. First, Phase I improves nominal robustness: under randomization, its average success remains stable from 44.6 to 44.0, whereas $\pi_0$ drops from 33.9 to 12.9. This indicates that recovery-aware initialization does not trade away base task competence. Second, under injected failures, full RePO-VLA improves over $\pi_{0.5}$ from 15.0/15.4 to 37.0/43.0. The gain is largest in tasks where the perturbation forces a genuine re-grasp or re-alignment, indicating that the combined RAI+VCR training helps the policy return to the success manifold rather than continuing the nominal plan. Baseline models such as RDT and GO-1 show lower success under randomized conditions, highlighting their limited robustness.

\subsection{Real-World Evaluation \& Recovery Data Scaling}
\label{sec:real_world_exp}

We evaluate on two Dobot X-Trainer arms across \textit{Vegetable Preparation}, \textit{Towel Folding}, \textit{Desk Organization}, and \textit{Liquid Pouring}. These tasks cover precision pouring, deformable-object manipulation, workspace rearrangement, and bimanual coordination. Each task uses 200 expert demonstrations and 50 teleoperated recovery episodes. In the adversarial setting, a human injects one physical perturbation per trial---dynamic displacement, kinematic interference, or forced slippage---at a pre-specified task stage, producing disturbances that are difficult to script exactly but common in physical deployments. Checkpoints are evaluated blind; the protocol is in \hyperref[sec:appendix_blind_protocol]{Appendix~\ref*{sec:appendix_blind_protocol}}, and qualitative task settings are shown in \hyperref[app:real_world_viz_sec]{Appendix~\ref*{app:real_world_viz_sec}}.



\begin{table}[t]
\centering
\caption{\textbf{FRBench-Real success rate (\%).} Results are averaged over 10 trials per task under standard and adversarial settings. Method labels include venue year and citation; $\Delta$ reports improvement over $\pi_{0.5}$ in the same condition.}
\label{tab:Real_world_combined}
\scriptsize
\setlength{\tabcolsep}{2.4pt}
\renewcommand{\arraystretch}{0.94}
\resizebox{\linewidth}{!}{
\begin{tabular}{lcccccccccc}
\toprule
\multirow{2}{*}{Task} & \multicolumn{5}{c}{Standard (No Perturbation) $\uparrow$} & \multicolumn{5}{c}{Adversarial (With Perturbation) $\uparrow$} \\
\cmidrule(lr){2-6} \cmidrule(lr){7-11}
& \piZeroTab & \piHalfTab & \cellcolor{repoRow}Phase I & \cellcolor{repoRow}Full (1x) & $\Delta$ & \piZeroTab & \piHalfTab & \cellcolor{repoRow}Phase I & \cellcolor{repoRow}Full (1x) & $\Delta$ \\
\midrule
Pour Water     & 40 & 30 & \cellcolor{repoRow}60 & \cellcolor{repoRow}50 & \gain{+20} & 20 & 20 & \cellcolor{repoRow}60 & \cellcolor{repoRow}30 & \gain{+10} \\
Cook Vegetable & 10 & 10 & \cellcolor{repoRow}30 & \cellcolor{repoRow}40 & \gain{+30} & 0  & 10 & \cellcolor{repoRow}30 & \cellcolor{repoRow}30 & \gain{+20} \\
Tidy Desk      & 20 & 20 & \cellcolor{repoRow}40 & \cellcolor{repoRow}40 & \gain{+20} & 10 & 30 & \cellcolor{repoRow}40 & \cellcolor{repoRow}30 & \gain{+0} \\
Fold Towel     & 40 & 40 & \cellcolor{repoRow}40 & \cellcolor{repoRow}30 & \loss{-10} & 20 & 20 & \cellcolor{repoRow}20 & \cellcolor{repoRow}30 & \gain{+10} \\
\cmidrule(lr){1-11}
Avg.           & 27.5 & 25.0 & \cellcolor{repoRow}\best{42.5} & \cellcolor{repoRow}\second{40.0} & \gain{+15.0} & 12.5 & 20.0 & \cellcolor{repoRow}\best{37.5} & \cellcolor{repoRow}\second{30.0} & \gain{+10.0} \\
\bottomrule
\end{tabular}}
\end{table}

%
\textbf{Baseline Fragility and Data Scaling.}
Table~\ref{tab:Real_world_combined} shows that imitation baselines degrade under adversarial perturbations. Phase I raises the adversarial average to 37.5\%. Full (1x) improves over $\pi_{0.5}$ but does not consistently outperform Phase I: VCR adds the full mixed trajectories, and at 1x the failure and recovery regions are too sparse to model reliably. We do not claim that VCR wins at every data scale.

\begin{table}[t]
\centering
\caption{\textbf{Recovery-data scaling study (\%).} We increase real-world recovery data from 1x to 2x and 4x on two representative tasks. Method labels include venue year and citation; $\Delta_{\mathrm{4x}-0.5}$ and $\Delta_{\mathrm{4x-I}}$ compare Full$^{**}$ (4x) with $\pi_{0.5}$ and Phase I.}
\label{tab:recovery_scaling}
\scriptsize
\setlength{\tabcolsep}{3.4pt}
\renewcommand{\arraystretch}{1.08}
\resizebox{\linewidth}{!}{
\begin{tabular}{llcccccccc}
\toprule
\multirow{2}{*}{Condition} & \multirow{2}{*}{Task} & \multicolumn{2}{c}{Baselines} & \multicolumn{4}{c}{RePO-VLA Variants} & \multicolumn{2}{c}{Improvement $\uparrow$} \\
\cmidrule(lr){3-4} \cmidrule(lr){5-8} \cmidrule(lr){9-10}
& & \piZeroTab & \piHalfTab & \cellcolor{repoRow}Phase I & \cellcolor{repoRow}Full (1x) & \cellcolor{repoRow}Full$^{*}$ (2x) & \cellcolor{repoRow}Full$^{**}$ (4x) & $\Delta_{\mathrm{4x}-0.5}$ & $\Delta_{\mathrm{4x-I}}$ \\
\midrule
\multirow{3}{*}{Standard} 
& Pour Water & 40 & 30 & \cellcolor{repoRow}60 & \cellcolor{repoRow}50 & \cellcolor{repoRow}\second{80} & \cellcolor{repoRow}\best{90} & \gain{+60} & \gain{+30} \\
& Fold Towel & 40 & 40 & \cellcolor{repoRow}40 & \cellcolor{repoRow}30 & \cellcolor{repoRow}\second{60} & \cellcolor{repoRow}\best{70} & \gain{+30} & \gain{+30} \\
\cmidrule(lr){2-10}
& Avg. & 40 & 35 & \cellcolor{repoRow}50 & \cellcolor{repoRow}40 & \cellcolor{repoRow}\second{70} & \cellcolor{repoRow}\best{80} & \gain{+45} & \gain{+30} \\
\midrule
\multirow{3}{*}{Adversarial}
& Pour Water & 20 & 20 & \cellcolor{repoRow}60 & \cellcolor{repoRow}30 & \cellcolor{repoRow}\second{70} & \cellcolor{repoRow}\best{80} & \gain{+60} & \gain{+20} \\
& Fold Towel & 20 & 20 & \cellcolor{repoRow}20 & \cellcolor{repoRow}30 & \cellcolor{repoRow}\second{60} & \cellcolor{repoRow}\best{70} & \gain{+50} & \gain{+50} \\
\cmidrule(lr){2-10}
& Avg. & 20 & 20 & \cellcolor{repoRow}40 & \cellcolor{repoRow}30 & \cellcolor{repoRow}\second{65} & \cellcolor{repoRow}\best{75} & \gain{+55} & \gain{+35} \\
\bottomrule
\end{tabular}}
\end{table}

Table~\ref{tab:recovery_scaling} reports the corresponding recovery-data scaling study. Doubling and quadrupling recovery data reduces the 1x bottleneck: Full$^{**}$ (4x) improves average standard and adversarial success over $\pi_{0.5}$ by 45 and 55 points, respectively, and surpasses Phase I under adversarial perturbations. This does not mean value conditioning wins at every scale.

\subsection{Ablation Studies}
\label{sec:ablations}

\begin{figure}[H]
    \centering
    \includegraphics[width=\linewidth]{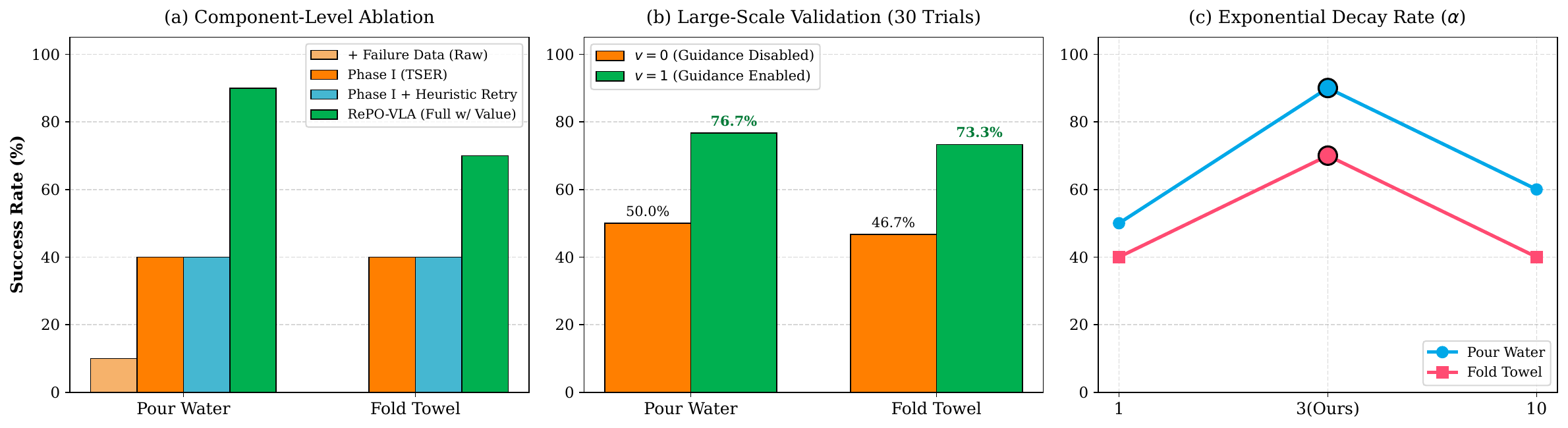}
    \caption{\textbf{Real-world ablations.} (a) Raw aggregation is worse than Phase I; see Table~\ref{tab:fold_towel_controls} for matched controls. (b) On Fold Towel, the high-value token instead of $v=0$ raises success from 14/30 (46.7\%) to 22/30 (73.3\%). (c) On Fold Towel, $\alpha=1,3,10$ yield 40\%, 70\%, and 40\%.}
    \label{fig:ablations}
\end{figure}

Figure~\ref{fig:ablations}(a) shows that directly adding raw failure-recovery trajectories performs substantially worse than Phase I, so imperfect data must be organized rather than appended uniformly. Table~\ref{tab:fold_towel_controls} gives the matched Fold Towel controls, including SARM+RA-BC \citep{chen2026sarm}. Table~\ref{tab:recap_offline} compares an RLinf \citep{yu2026rlinf} reproduction of offline RECAP with matched steps, recovery budget, and blind protocol. Details are in \hyperref[sec:appendix_added_controls]{Appendix~\ref*{sec:appendix_added_controls}}.

\begin{table}[t]
\centering
\caption{\textbf{Fold Towel controls (\%).} $\Delta$ is versus Success-only BC. ``--'' was not evaluated. Full-4x without value and RePO-VLA Full-4x share initialization, data, budget, and protocol.}
\label{tab:fold_towel_controls}
\scriptsize
\setlength{\tabcolsep}{2.2pt}
\renewcommand{\arraystretch}{0.92}
\begin{tabular*}{\linewidth}{@{\extracolsep{\fill}}lcccccc@{}}
\toprule
\multirow{2}{*}{Setting} & \multirow{2}{*}{Data} & \multirow{2}{*}{Value} & \multicolumn{2}{c}{Standard $\uparrow$} & \multicolumn{2}{c}{Adversarial $\uparrow$} \\
\cmidrule(lr){4-5} \cmidrule(lr){6-7}
& & & Rate & $\Delta$ & Rate & $\Delta$ \\
\midrule
Success-only BC & None & No & 40 & \na & 20 & \na \\
Raw DAgger-style & Raw & No & 0 & \loss{-40} & -- & -- \\
Phase-I DAgger-style & Sliced + reset & No & 40 & \gain{+0} & 20 & \gain{+0} \\
Full-4x, no value & Full 4x & No & 20 & \loss{-20} & 20 & \gain{+0} \\
SARM+RA-BC, Full-4x & Full 4x & External & \second{50} & \gain{+10} & 20 & \gain{+0} \\
\rowcolor{repoRow}
RePO-VLA Full-4x & Full 4x & Yes & \best{70} & \gain{+30} & \best{70} & \gain{+50} \\
\bottomrule
\end{tabular*}
\end{table}

\begin{table}[t]
\centering
\caption{\textbf{Offline RECAP comparison (\%).} RECAP-1x ($N{=}50$) and RECAP-2x ($N{=}100$) use the same $\pi_{0.5}$ backbone. Bold/underline rank RECAP-1x, RECAP-2x, and Full-2x. $\Delta$ is Full-2x minus RECAP-2x. Only the offline advantage-conditioned component is compared.}
\label{tab:recap_offline}
\scriptsize
\setlength{\tabcolsep}{2.2pt}
\renewcommand{\arraystretch}{0.92}
\begin{tabular*}{\linewidth}{@{\extracolsep{\fill}}llccccc@{}}
\toprule
\multirow{2}{*}{Condition} & \multirow{2}{*}{Task} & \multicolumn{3}{c}{Baseline and RECAP} & \multicolumn{2}{c}{RePO-VLA} \\
\cmidrule(lr){3-5} \cmidrule(lr){6-7}
& & \makecell[c]{$\pi_{0.5}$\\[-1pt]{\tiny CoRL'25}} & \makecell[c]{RECAP-1x\\[-1pt]{\tiny $N{=}50$}} & \makecell[c]{RECAP-2x\\[-1pt]{\tiny $N{=}100$}} & \cellcolor{repoRow}\makecell[c]{Full-2x\\[-1pt]{\tiny RePO-VLA}} & $\Delta$ \\
\midrule
\multirow{3}{*}{Standard}
& Pour Water & 30 & \second{60} & \best{80} & \cellcolor{repoRow}\best{80} & \gain{+0} \\
& Fold Towel & 40 & 50 & \best{70} & \cellcolor{repoRow}\second{60} & \loss{-10} \\
\cmidrule(lr){2-7}
& Average & 35 & 55 & \best{75} & \cellcolor{repoRow}\second{70} & \loss{-5} \\
\midrule
\multirow{3}{*}{Adversarial}
& Pour Water & 20 & \second{40} & \best{70} & \cellcolor{repoRow}\best{70} & \gain{+0} \\
& Fold Towel & 20 & 20 & \second{50} & \cellcolor{repoRow}\best{60} & \gain{+10} \\
\cmidrule(lr){2-7}
& Average & 20 & 30 & \second{60} & \cellcolor{repoRow}\best{65} & \gain{+5} \\
\bottomrule
\end{tabular*}
\end{table}

\textbf{Decay Rate.}
Figure~\ref{fig:ablations}(c) identifies $\alpha=3$ as the best trade-off. A small $\alpha$ insufficiently penalizes near-terminal failure actions, whereas a large $\alpha$ suppresses useful pre-failure behavior.

\vspace{-2mm}

\vspace{-2mm}
\section{Conclusion}
We presented RePO-VLA, a recovery-driven framework for training VLA policies from nominal-success, failed, and recovered trajectories. By combining episode-reset recovery initialization with progress-aware value conditioning, RePO-VLA converts mixed-quality interaction data into autonomous corrective behavior. The central result is that failures need not be discarded: their useful pre-failure segments can provide kinematic priors, while their terminal regions can be safely down-weighted through semantic progress labels. Results on the Failure-Recovery Benchmark (FRBench) show that explicit recovery supervision and dense value guidance are key ingredients for reliable long-horizon bimanual manipulation. The method supplies this supervision during fine-tuning and does not replace the pretrained backbone.

\vspace{-2mm}
\section{Limitations and Future Work}
RePO-VLA is recovery-aware optimization on observed failure families, not open-world zero-shot recovery. FRBench-Sim covers premature closure, grasp slip, position offset, and orientation mismatch. FRBench-Real adds deformable and fluid interaction, dynamic displacement, forced slippage, and bimanual kinematic interference. Severe occlusion, tool misuse, subgoal mismatch, and major perception errors are outside this evaluation. Teleoperated recovery limits scale; the planner and the operator are used only offline. Reducing that annotation, and modeling failure inside pretraining, are left as future work.

\begin{ack}
This work was supported by the National Key Research and Development Program
of China (Grant No.\ 2024YFE0203100); the Scientific Research Innovation Capability
Support Project for Young Faculty (Grant No.\ ZYGXQNJSKYCXNLZCXM-I28); the
National Natural Science Foundation of China (NSFC) (Grant Nos.\ 62476293,
62636010, and U24A20329); the Shenzhen Science and Technology Program
(Grant No.\ QNXMA20250701093544048); the Guangdong Basic and Applied Basic
Research Foundation (Grant No.\ 2024B1515120024); the General Embodied AI Center
of Sun Yat-sen University; and the Major Key Project of Pengcheng Laboratory
(Grant No.\ PCL2025A17).

\textbf{Competing interests.} The authors declare no financial relationships
outside the submitted work during the 36 months prior to submission that
could be perceived as influencing this work.
\end{ack}

{
\small




\bibliographystyle{plainnat} 
\bibliography{reference}    
}




\clearpage
\appendix
\section{Appendix}
\label{sec:appendix}
\subsection{Roadmap and Acronyms}
\label{sec:appendix_roadmap}

\begingroup
\large
The appendix is organized as follows; each entry links to the corresponding section.
\begin{itemize}[leftmargin=*, topsep=4pt, itemsep=3pt, parsep=0pt, partopsep=0pt]
    \item \hyperref[sec:appendix_repo]{Appendix~\ref*{sec:appendix_repo}: Additional Details on RePO-VLA} covers episode reset, the RAI/VCR data split, detector-free inference, reliability decay, and compute resources.
    \item \hyperref[sec:appendix_more_related_work]{Appendix~\ref*{sec:appendix_more_related_work}: More Related Work} expands the compact discussion in the main paper.
    \item \hyperref[sec:appendix_frbench]{Appendix~\ref*{sec:appendix_frbench}: Additional Details on FRBench} specifies the protocol, perturbation taxonomy, detection logic, and dataset composition.
    \item \hyperref[sec:appendix_added_controls]{Appendix~\ref*{sec:appendix_added_controls}: Added Real-World Controls} explains the DAgger-style procedure, the SARM progress diagnostic, the RECAP data distinction, and the blind adversarial protocol. The success rates are in Section~\ref{sec:ablations}.
    \item \hyperref[app:data_viz]{Appendix~\ref*{app:data_viz}: Visualization of Recovery Data} shows simulated failure-recovery episodes.
    \item \hyperref[app:real_world_viz_sec]{Appendix~\ref*{app:real_world_viz_sec}: Qualitative Real-World Evaluation} shows the real-world task settings.
\end{itemize}
\endgroup
Table~\ref{tab:acronyms} summarizes recurring abbreviations and notation.

\begin{table}[!htbp]
\centering
\caption{\textbf{Acronym and notation reference.}}
\label{tab:acronyms}
\scriptsize
\setlength{\tabcolsep}{3pt}
\renewcommand{\arraystretch}{1.04}
\begin{tabularx}{\linewidth}{@{}>{\raggedright\arraybackslash}p{0.15\linewidth}>{\raggedright\arraybackslash}p{0.27\linewidth}X@{}}
\toprule
Term & Full name & Meaning in this paper \\
\midrule
VLA & Vision-Language-Action & Robot policy family that maps visual observations and language instructions to actions. \\
RePO-VLA & Recovery-driven Policy Optimization for VLA & Our training framework for autonomous recovery from physical execution drift. \\
FRBench & Failure-Recovery Benchmark & Benchmark and data engine for evaluating recovery after controlled adverse-state projection. \\
RAI & Recovery-Aware Initialization & First training stage. It injects recovery priors from expert demonstrations and sliced corrective episodes. \\
TSER & Trajectory Slicing with Episode Reset & RAI preprocessing that drops the failure prefix and restarts the corrective suffix as a new episode at $t=0$. \\
PAS-VF & Progress-Aware Semantic Value Function & Dense progress estimator based on semantic alignment to successful trajectory features. \\
VCR & Value-Conditioned Refinement & Second training stage. It fine-tunes the policy on mixed trajectories with dense value labels. \\
SFT & Supervised Fine-Tuning & Imitation-style fine-tuning on action-labeled demonstrations. \\
HITL / MITL & Human-/Model-in-the-Loop & External correction or replanning mechanisms used by prior recovery systems. \\
RLFT & Reinforcement Fine-Tuning & Policy refinement using reward or preference feedback. \\
V-JEPA & Video Joint-Embedding Predictive Architecture & Frozen spatiotemporal encoder used by PAS-VF. \\
E1--E4 & Error types 1--4 & FRBench perturbation taxonomy: premature close, grasp slip, position offset, and orientation mismatch. \\
\bottomrule
\end{tabularx}
\end{table}

\clearpage

\subsection{Additional Details on RePO-VLA}
\label{sec:appendix_repo}

\subsubsection{Why Episode Reset Matters}
\label{sec:appendix_history_reset}
Recovery demonstrations naturally contain a failure prefix followed by corrective actions. The base policy reads a single-timestamp observation and does not stack historical image frames, so ``reset'' does not clear an observation buffer. The risk is at the episode level: if the prefix and the suffix stay in one training episode, the policy can associate corrective actions with the preceding mistake. TSER removes that shortcut during RAI by discarding the failure prefix and writing the corrective suffix as a new episode that begins at $t=0$. The model therefore infers the correction from the present observation, such as an empty gripper or a displaced object.

\subsubsection{Episode Boundaries Across RAI, VCR, and Inference}
\label{sec:appendix_history_treatment}
RAI and VCR use different episode boundaries, not different observation-history buffers. RAI trains on $\mathcal{D}_{rec}^{reset}$, where each corrective suffix is its own episode, so the failure prefix cannot act as a trigger. VCR returns to $\mathcal{D}_{rec}^{raw}$ and keeps the prefix and the suffix in one episode, using low value on the failure-inducing actions and high value on the corrective actions. At deployment there is no episode reset and no online value estimate: the policy sees the current observation and the fixed high-value token $v=1.0$.

\subsubsection{Inference Without Online Failure Detection}
\label{sec:appendix_no_detector}
RePO-VLA decouples value estimation during training from deployment-time recovery. At test time, we do not estimate $V(\tau)$ online or trigger a recovery module. The policy is always conditioned on the high-value token $v=1.0$, the normalized success anchor of our labels. This makes high-progress behavior the default target; when drift occurs, the same policy selects actions that move the state back toward the learned success manifold.

\subsubsection{Reliability Decay for Pure Failures}
\label{sec:appendix_decay}
Pure failures often contain useful approach behavior before the terminal breakdown. Assigning zero to every frame discards these kinematic priors, while assigning high value near failure rewards harmful actions. The decay label
$v_t=V(\tau_{fail})(1-t/T)^\alpha$
keeps early useful behavior and rapidly downweights late failure states. We use $\alpha=3.0$ in all main experiments.

\subsubsection{Compute Resources}
\label{sec:appendix_compute}
All training and data-generation jobs were run on an internal server with two NVIDIA A100 GPUs, each with 80 GB of GPU memory, 128 CPU cores, and 512 GB of host RAM. RAI and VCR used both A100 GPUs and required approximately 30 hours per full training run, including value-label preparation and PAS-VF adapter training. PAS-VF trains only lightweight adapters on top of frozen encoders; the exact optimizer, batch size, learning rate, and step count follow the released training configuration. FRBench-Sim data generation in RoboTwin used up to 128 CPU workers on the same server, with GPU resources used for policy/planner inference and visual feature extraction. The real-world evaluation used two Dobot X-Trainer arms and did not require online policy optimization; deployment only required policy inference on a single GPU.

\subsection{More Related Work}
\label{sec:appendix_more_related_work}

This section expands the compact related work discussion in the main paper while keeping the central distinction unchanged: RePO-VLA is not another VLA backbone or external retry module, but a way to reuse success, failure, and recovery data under progress-aligned value supervision.

\paragraph{VLA backbones and action modeling.}
Recent VLA models have rapidly improved the base policy interface. OpenVLA \citep{kim2024openvla} casts actions as language-like tokens, RDT-1B \citep{liu2024rdt} scales diffusion modeling for bimanual manipulation, and the $\pi$ series \citep{black2024pi_0, intelligence2025pi_} uses flow matching to support continuous robot control. DiffusionVLA \citep{wen2025diffusionvla} and OTTER \citep{huang2025otter} further study unified diffusion/autoregressive training and text-aware visual extraction. These works strengthen nominal policy capacity, but they do not specify how failed rollouts should be decomposed into useful prefixes, low-value drift, and corrective recovery.

\paragraph{Failure recovery and runtime correction.}
Recovery-oriented systems often add an external reasoning or correction layer. ReplanVLM \citep{mei2024replanvlm}, RACER \citep{dai2025racer}, and RoboFAC \citep{lu2025robofac} use visual-language reasoning or failure analysis to produce recovery plans, while FailSafe \citep{lin2025failsafe} synthesizes failure-recovery data for VLA training. RoboMD \citep{sagar2024robomd} studies policy vulnerabilities and motivates explicit failure-aware evaluation. RePO-VLA is complementary to these efforts, but differs by internalizing recovery into the low-level policy through dense value labels rather than relying on a separate detector, planner, or uniformly positive recovery imitation.

\paragraph{Learning from mixed-quality robotic data.}
Offline-to-online RL and preference-style refinement provide useful tools for reusing non-ideal data. RLPD \citep{ball2023efficient} improves online RL with offline demonstrations, ConRFT \citep{chen2025conrft} aligns VLA policies through consistency-based reinforced fine-tuning, ReinboT \citep{zhang2025reinbot} leverages dense returns for visual-language manipulation, and WMPO \citep{zhu2025wmpo} optimizes VLA behavior with learned world-model feedback. RePO-VLA targets a more specific failure-recovery structure: it gives different semantic roles to successful segments, failure prefixes, terminal drift, and corrective actions, allowing the policy to learn distributional boundaries between harmful and reparative behavior.

\subsection{Additional Details on FRBench} \label{sec:appendix_frbench}

This section details the FRBench failure-recovery data engine, including the phase-based protocol, synthetic error injection mechanism, autonomous detection logic, and dataset scale.

\subsubsection{Phase-Based Evaluation Protocol Details}
\label{sec:appendix_frbench_protocol}
FRBench logs a phase variable $\phi_t\in\{\textsc{Nominal},\textsc{Error},\textsc{Recovery}\}$ for every evaluation trajectory. A rollout begins in $\textsc{Nominal}$, where the policy is evaluated under the same task specification as a standard success benchmark. At a task-specific critical node $t_{crit}$, the benchmark enters $\textsc{Error}$ and applies a deterministic perturbation operator from the E1--E4 taxonomy. Recovery evaluation begins only after the adverse state is verified, so the recovery score is conditional on a physical breakdown having occurred rather than on the policy merely reaching the perturbation point. During $\textsc{Recovery}$, the policy continues from the perturbed state and must satisfy the task's recovery condition without an external retry controller. This phase separation prevents nominal improvements, such as better reaching or grasp timing, from being mistaken for genuine corrective behavior.

\subsubsection{The 5-Step Interception Sequence}
\label{sec:appendix_frbench_sequence}
The synthetic error injection follows a strictly controlled execution flow that intercepts the native planner without altering the base pipeline. To autonomously generate paired failure-recovery episodes at scale, the generation process follows a strict five-step sequence:

\begin{enumerate}[leftmargin=*, label=\textbf{Step \arabic*:}, parsep=2pt]
    \item \textbf{Nominal Reference Generation:} The expert oracle computes a complete, successful trajectory $\tau_{nom}$ for a given task specification.
    \item \textbf{Execution Hooking:} We continuously monitor the task progression and intercept the critical execution node, such as the exact moment of a grasp or lift motion ($t_{crit}$).
    \item \textbf{Perturbation Injection:} At $t_{crit}$, we inject a targeted error type (E1--E4) by overriding the oracle's standard control output $a_t$ with predefined perturbation operators.
    \item \textbf{Adverse State Projection:} The robot executes the perturbed trajectory, deliberately inducing a natural physical breakdown. This reliably projects the system into a deterministic "adverse state."
    \item \textbf{Hierarchical Recovery:} Once the failure is registered, the hierarchical recovery planner is activated to dynamically replan and synthesize the corrective actions needed to restore the task state.
\end{enumerate}

\subsubsection{Mathematical Formulation of Perturbation Taxonomy (E1--E4)}
\label{sec:appendix_frbench_taxonomy}
A core contribution of FRBench is the parameterization of physical breakdowns. Let the nominal base action at timestep $t$ be defined as $a_t = [p_t, R_t, g_t]$, where $p_t \in \mathbb{R}^3$ is the end-effector position, $R_t \in SO(3)$ is the orientation, and $g_t$ represents the continuous gripper state. To guarantee reproducible execution drift without relying on stochastic physics engine artifacts, the four error types are implemented as direct control overrides:

\begin{itemize}[leftmargin=*, topsep=4pt, itemsep=4pt, parsep=0pt, partopsep=0pt]
    \item \textbf{E1: \texttt{premature\_close}.} The gripper closes early during the approach stage of a grasp primitive, often resulting in an empty grasp or partial contact. Mathematically, we override the gripper command to a closed state prior to reaching the target: $g'_t = g_{closed} \quad \text{for } t \in [t_{approach}, t_{approach} + N_{hold}]$.
    \item \textbf{E2: \texttt{grasp\_slip}.} The object is released during the lift phase to simulate insufficient grasp force, causing a drop or loss of grasp. We model this dynamic slippage proxy by forcefully opening the gripper for a short, deterministic window to abruptly break contact forces: $g'_t = g_{open} \quad \text{for } t \in [t_{lift}, t_{lift} + 30]$. 
    \item \textbf{E3: \texttt{grasp\_position\_offset}.} A cm-level translational offset is applied to the grasp target pose, modeling visual misalignment and localization bias. The perturbed position is defined as $p'_t = p_t + \delta p_{trans}$, where $\delta p_{trans} \sim \mathcal{U}(-d, d)$, forcing the gripper to close at an offset.
    \item \textbf{E4: \texttt{grasp\_orientation\_mismatch}.} A large orientation mismatch, coupled with a stabilizing lateral shift, is applied to induce a realistic misgrasp without trivial physical collisions. The rotational override is formulated as $R'_t = R_t \cdot \Delta R_{err}$, accompanied by an adjustment $p'_t = p_t + \delta p_{lat}$.
\end{itemize}
The E1--E4 operators define controlled breakdowns in FRBench-Sim. Real-world adversarial trials do not replay these scripted gripper overrides. Each real trial applies one physical perturbation---dynamic displacement, forced slippage, or bimanual kinematic interference---at a pre-specified task stage, as detailed in \hyperref[sec:appendix_blind_protocol]{Appendix~\ref*{sec:appendix_blind_protocol}}.

\subsubsection{Detailed Detection and Recovery Logic}
\label{sec:appendix_frbench_detection}
\textbf{Autonomous Failure Detection:} In simulation, relying solely on collision or object-drop heuristics is insufficient for complex tasks. Therefore, failure events $\mathcal{F}_{t}$ are triggered when the execution horizon exceeds a task-specific threshold defined by the maximum nominal success duration: 
\begin{equation}
    \mathcal{F}_{t} = \mathbb{I}(t > T_{max}), \quad \text{where } T_{max} = \max_{i \in \{1,\dots,n\}}(T_{succ}^i)
\end{equation}
This temporal heuristic allows the data engine to autonomously capture insidious failure modes—such as execution loops or progress stagnation—that do not trigger immediate physical collisions but definitively signify task failure.

\textbf{Real-World Recovery Protocol:} Unlike the automated simulation planner, real-world recovery episodes are curated via human-in-the-loop (HITL) teleoperation. Operators intentionally drive the robot into "Adverse States" (e.g., an object partially slipped, or inter-arm kinematic misalignment) and provide a continuous corrective demonstration back to the nominal success manifold. This ensures that the high-value recovery behaviors ($\tau_{rec}^+$) are strictly grounded in real-world friction and contact dynamics, which are often unmodeled in simulation environments.

\subsubsection{Dataset Statistics and Composition}
\label{sec:appendix_frbench_stats}
The FRBench-Sim dataset comprises a large-scale collection of bimanual manipulation trajectories. It is generated via our automated pipeline within the RoboTwin simulation platform, utilizing the ALOHA-Agilex embodiment. In total, the dataset contains \textbf{23,453 episodes}: \textbf{17,061} nominal-success trajectories with no injected error, and \textbf{6,392} verified failure-recovery trajectories. Nominal episodes are unperturbed. Each retained failure-recovery episode contains one injected error. The E1--E4 counts below are injection attempts recorded before this filter, not a second inventory of the 23,453 episodes. Those attempts sum to 16,912, of which 6,392 were kept as verified recovery episodes.

\begin{table}[t]
\centering
\caption{\textbf{FRBench-Sim dataset statistics.} Counts summarize the generated trajectory corpus and the aggregate error-injection coverage.}
\label{tab:frbench_dataset_stats}
\scriptsize
\setlength{\tabcolsep}{6pt}
\renewcommand{\arraystretch}{1.08}
\begin{tabular}{lrl}
\toprule
Category & Count & Description \\
\midrule
Total episodes & 23,453 & All generated bimanual manipulation episodes \\
Nominal success & 17,061 & Unperturbed successful trajectories, with no injected error \\
Failure-recovery & 6,392 & Verified recovery episodes retained after filtering; each has one injected error \\
Task diversity & 46 & Distinct bimanual tasks across daily and industrial settings \\
Environment modes & 2 & Clean and randomized task instantiations \\
\midrule
E1: Premature Close & 8,022 & Early gripper closure during approach; pre-filter attempts \\
E2: Grasp Slip & 3,516 & Forced gripper opening during lift; pre-filter attempts \\
E3: Position Offset & 4,686 & Translational grasp-target offset; pre-filter attempts \\
E4: Orientation Mismatch & 688 & Rotational mismatch with a lateral shift; pre-filter attempts \\
\bottomrule
\end{tabular}
\vspace{1mm}

{\scriptsize \emph{Note.} The four error-type rows sum to 16,912 pre-filter injection attempts. They are not added on top of the 23,453 episodes. Of those attempts, 6,392 were retained as verified failure-recovery episodes. The other 17,061 episodes are nominal successes and were not injected.}
\end{table}

\subsection{Added Real-World Controls}
\label{sec:appendix_added_controls}

\subsubsection{DAgger-Style Aggregation}
Table~\ref{tab:fold_towel_controls} reports the Fold Towel controls in the main paper. The DAgger-style setting rolls out the policy to collect adverse states, obtains corrective trajectories from an expert planner or a human operator, aggregates them with the existing dataset, and then performs supervised fine-tuning. It is called DAgger-style rather than canonical DAgger because the expert provides a corrective trajectory suffix after an adverse state, instead of relabeling every visited state with a single expert action. Raw DAgger-style aggregation keeps the complete failure-recovery trajectory, including the failure-inducing prefix and the corrective suffix. Phase-I DAgger-style keeps only the expert corrective suffix and treats it as an independent episode with a reset boundary. The failure-inducing prefix should not be treated as expert supervision. Recovery data provides coverage of adverse states and corrective behaviors, while value-conditioned refinement helps the policy use this mixed-quality data. This is consistent with the scaling result in Table~\ref{tab:recovery_scaling}, where RePO-VLA on Fold Towel improves from 30\% at 1x to 60\% at 2x and 70\% at 4x.

\subsubsection{SARM+RA-BC}
SARM+RA-BC is an alternative learned-progress baseline from one stage-aware reward model \citep{chen2026sarm}. It does not use the PAS-VF labels or the temporal reliability-decay formulation. SARM estimates stage-aware progress. RA-BC is the behavior-cloning rule in that same model: it uses the progress difference between adjacent action chunks to filter or reweight the imitation loss. On Fold Towel, uniform Full-4x fine-tuning without value reaches 20\% standard and 20\% adversarial success. SARM+RA-BC reaches 50\% and 20\%. RePO-VLA Full-4x reaches 70\% and 70\%. The strict matched pair is uniform Full-4x without value versus RePO-VLA Full-4x. SARM+RA-BC is the external value-model baseline.

Learned progress is useful for mixed-quality data under the standard setting, but in this experiment it does not add robustness under forced perturbations. After a physical failure, the scene may still appear partially complete, so a progress estimator can output a moderate or plateaued value. Using that value directly would assign positive utility to actions near the failed endpoint. Reliability decay converts estimated progress into training utility: it preserves useful early motion while reducing the weight of actions approaching terminal failure.

On held-out Dobot failure-recovery rollouts, SARM generally captures the overall task order, but it occasionally shows stage mismatch, oscillation during re-grasping or backtracking, and spurious progress increases after a physical failure. Those errors can cause RA-BC to retain failure-inducing chunks or suppress corrective chunks. This observation is a qualitative diagnostic. The 50\%/20\% standard and adversarial results above are the policy-level evidence.

\subsubsection{Offline RECAP Comparison}
Table~\ref{tab:recap_offline} in the main paper reports the offline comparison. Official RECAP code and checkpoints were not available at submission. During the discussion period, the open-source RLinf \citep{yu2026rlinf} reproduction was ported to the Dobot platform. The comparison uses the same $\pi_{0.5}$ backbone, matched training steps, equal recovery-data budgets, and the same blind evaluation protocol. Because the complete RECAP system includes an autonomous online improvement loop, the experiment compares its advantage-conditioned learning component under a fixed offline budget. RECAP-1x is labeled $N=50$ and RECAP-2x is labeled $N=100$. RECAP collects experience from $\pi_{0.5}$ rollouts, where severe failures occur less frequently than in the targeted recovery-data collection, so its dataset contains a larger proportion of nominal and near-nominal experience. RePO-VLA deliberately induces adverse states and collects corrections from them, producing denser coverage of forced slips, displacement, and recovery states. The two results are complementary. Neither is evidence that one method dominates the other.

\subsubsection{Blind Adversarial Protocol}
\label{sec:appendix_blind_protocol}
The real-world evaluation is blind. Each checkpoint is served through a separate \texttt{ip:port}, and the port-to-method mapping is held by an author who does not participate in the trials. The operator selects only a port and does not know the method identity. All methods use the same scene, initial poses, control frequency, success criteria, and 10 trials per task and condition.

Each adversarial trial receives exactly one physical perturbation. Its type, nominal magnitude, and task-phase trigger are fixed by a pre-specified schedule, and the operator does not intervene again. Triggering by semantic phase rather than wall-clock time places the perturbation at the same task stage for policies with different execution speeds. All checkpoints were selected with the same pre-specified validation criterion and frozen before adversarial testing. No adversarial-test outcome was used for checkpoint selection.

\subsection{Visualization of Recovery Data}
\label{app:data_viz}

This appendix visualizes representative data samples generated by our automated pipeline within the RoboTwin simulation platform, utilizing the ALOHA-Agilex embodiment. The purpose is to intuitively illustrate the complete workflow of our system, specifically the physics-driven error injection and the subsequent autonomous recovery process.

For each of the four core error types (E1–E4), we provide six task examples across both Clean and Random (domain-randomized) environments to ensure visual and kinematic diversity. Each task sequence consists of 9 keyframes sampled from a single episode:

\begin{itemize}[leftmargin=*, topsep=0pt, itemsep=0pt, parsep=0pt, partopsep=0pt]
\item\texttt{Frames 1-3 (Error):}  Visualize the ``Error Attempt'' phase, where physics-driven perturbations (rather than simple Gaussian noise) are introduced to cause a natural failure.
\item\texttt{Frames 4-8 (Recovery):} Display the ``Recovery'' phase, featuring smooth, planner-based trajectories that include re-perception and corrective actions.
\item\texttt{Frame 9 (Normal):} Illustrate the robot successfully resuming the "Normal" execution phase from a corrected state.
\end{itemize}

\begin{figure*}[htbp]
    \centering
    \includegraphics[width=1.0\linewidth]{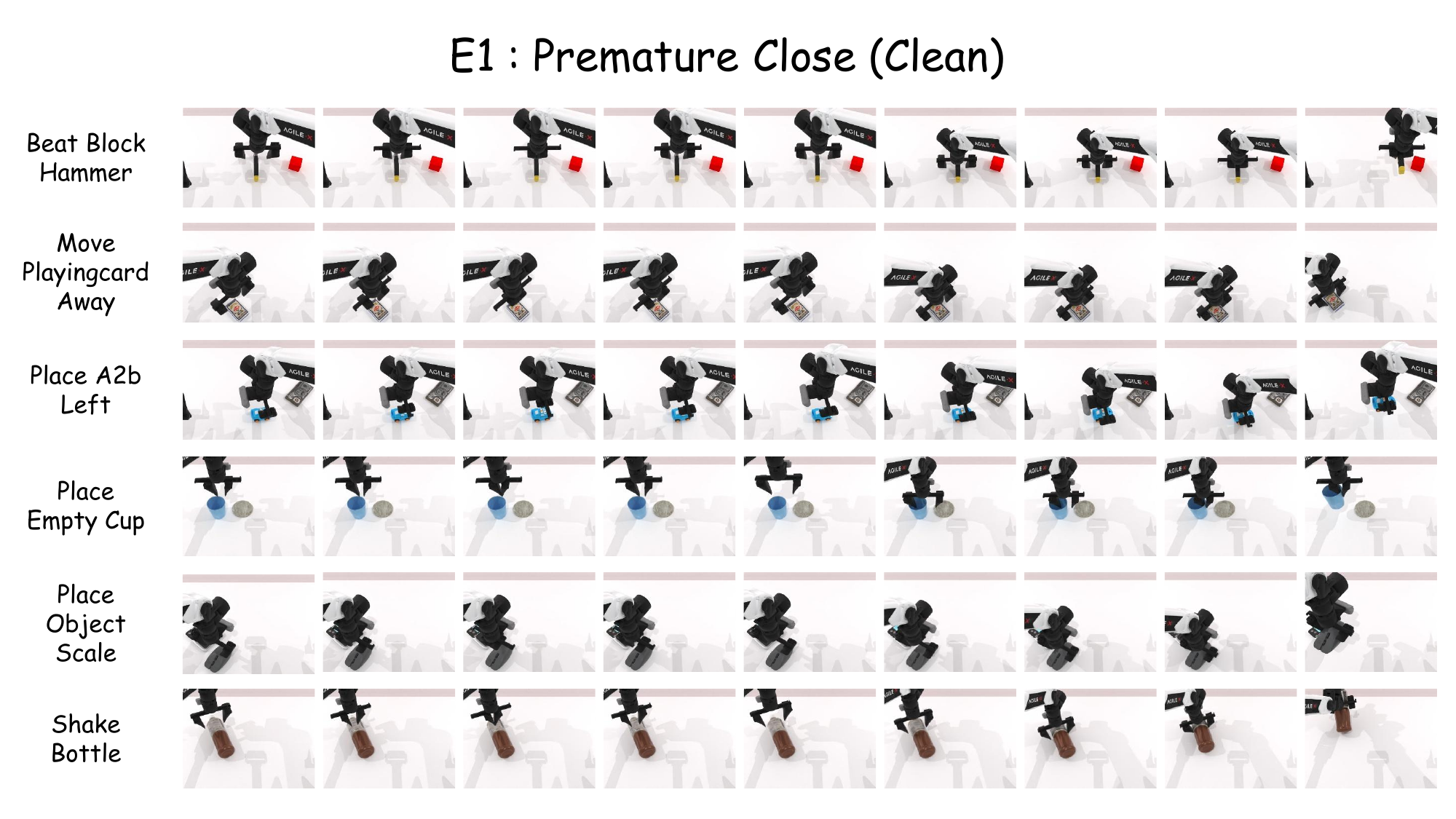}
    \label{fig:appendix_1}
    \caption{Premature Close (Clean)}
\end{figure*}

\begin{figure*}[htb]
    \centering
    \includegraphics[width=1.0\linewidth]{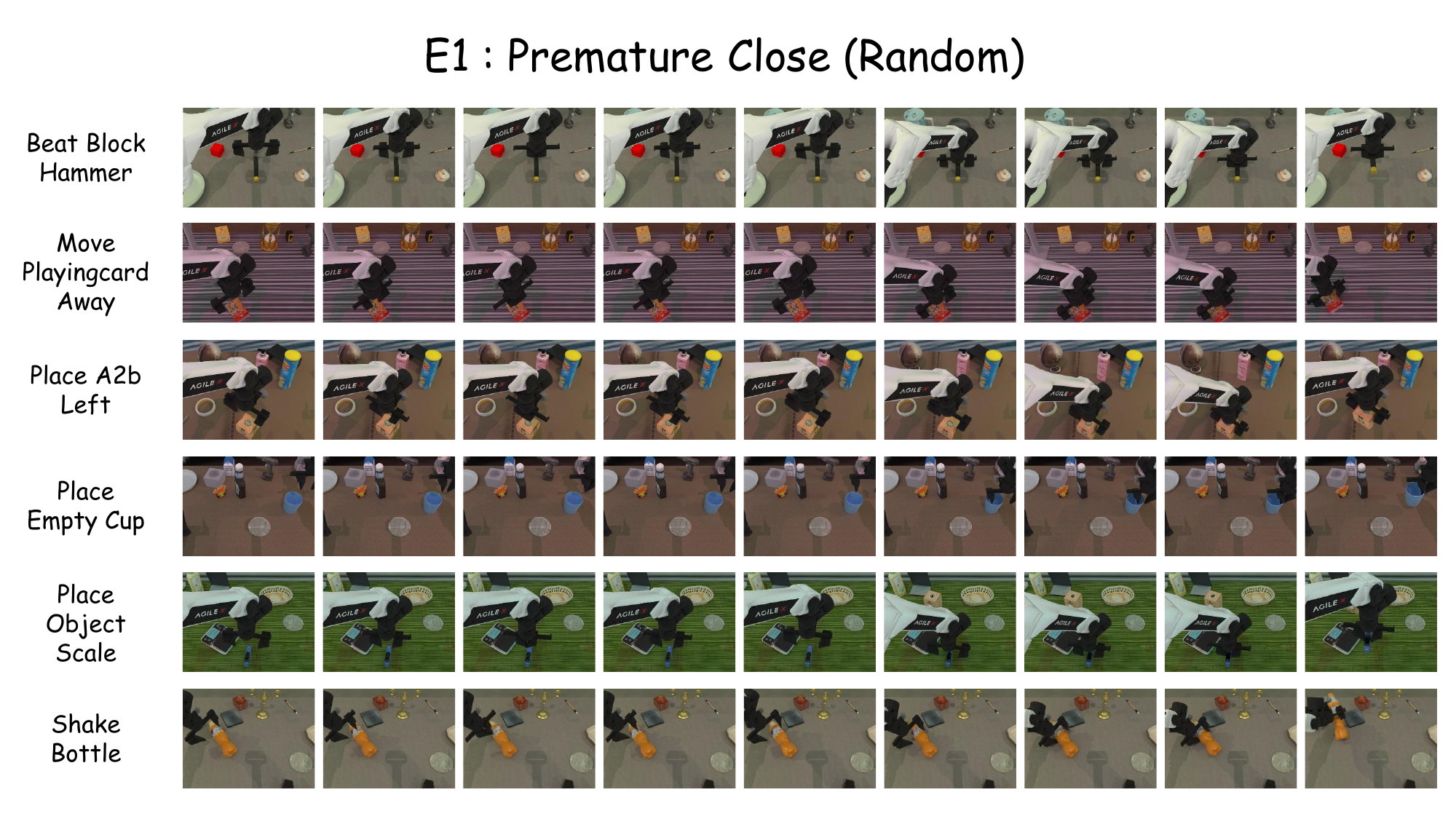}
    \label{fig:appendix_2}
    \caption{Premature Close (Random)}
\end{figure*}

\begin{figure*}[htb]
    \centering
    \includegraphics[width=1.0\linewidth]{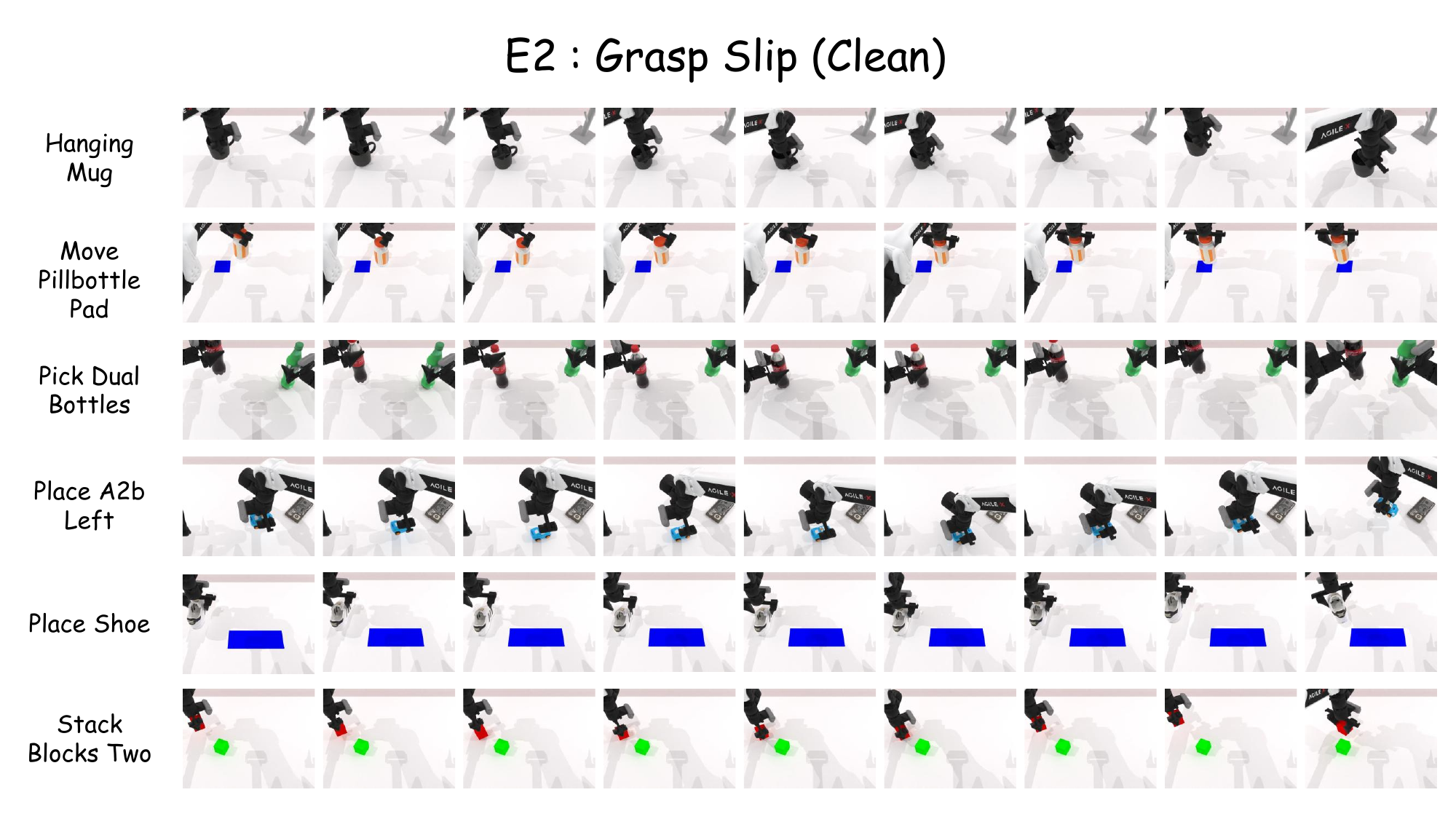}
    \label{fig:appendix_3}
    \caption{Grasp Slip (Clean)}
\end{figure*}

\begin{figure*}[htb]
    \centering
    \includegraphics[width=1.0\linewidth]{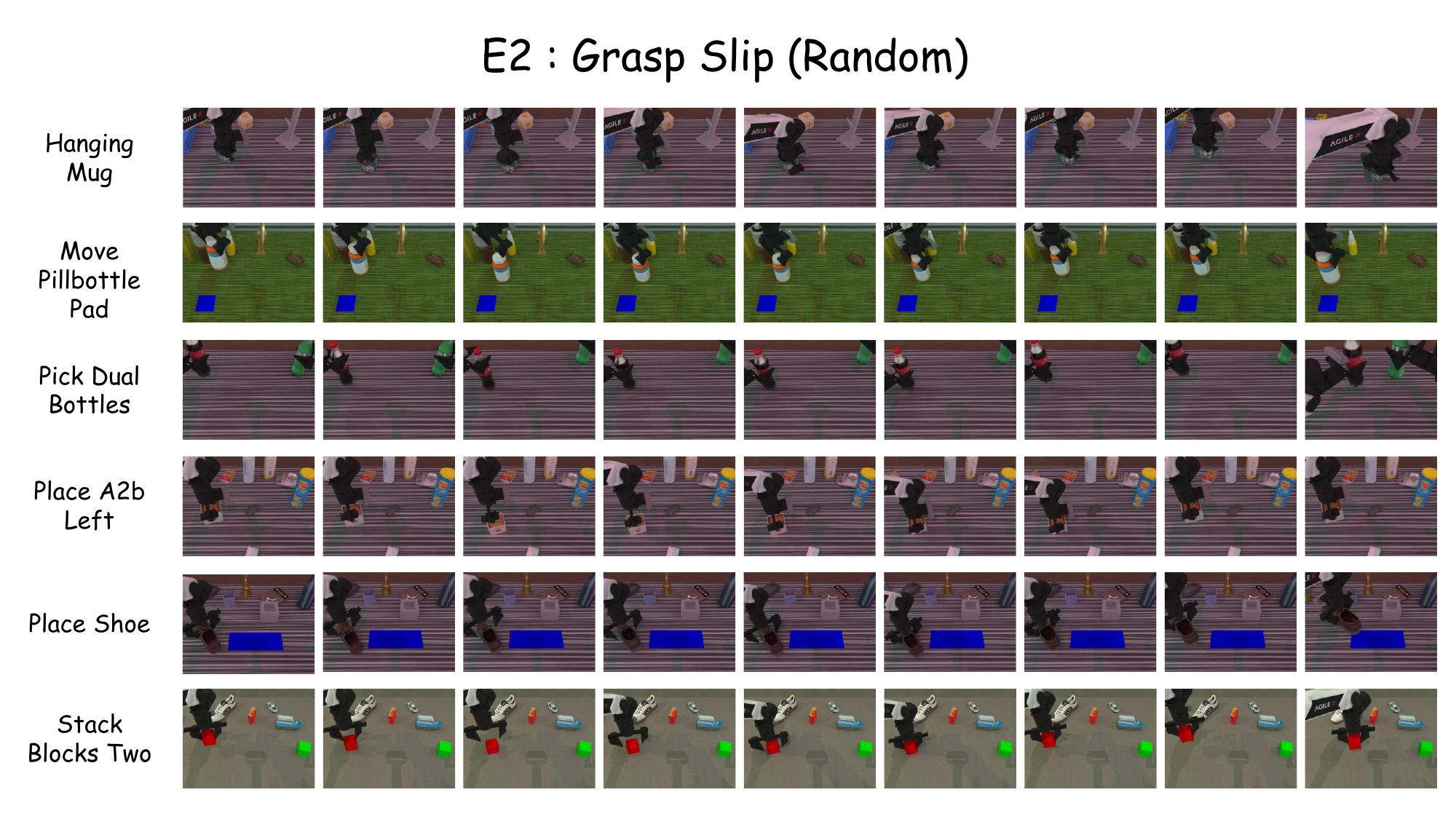}
    \label{fig:appendix_4}
    \caption{Grasp Slip (Random)}
\end{figure*}

\begin{figure*}[htb]
    \centering
    \includegraphics[width=1.0\linewidth]{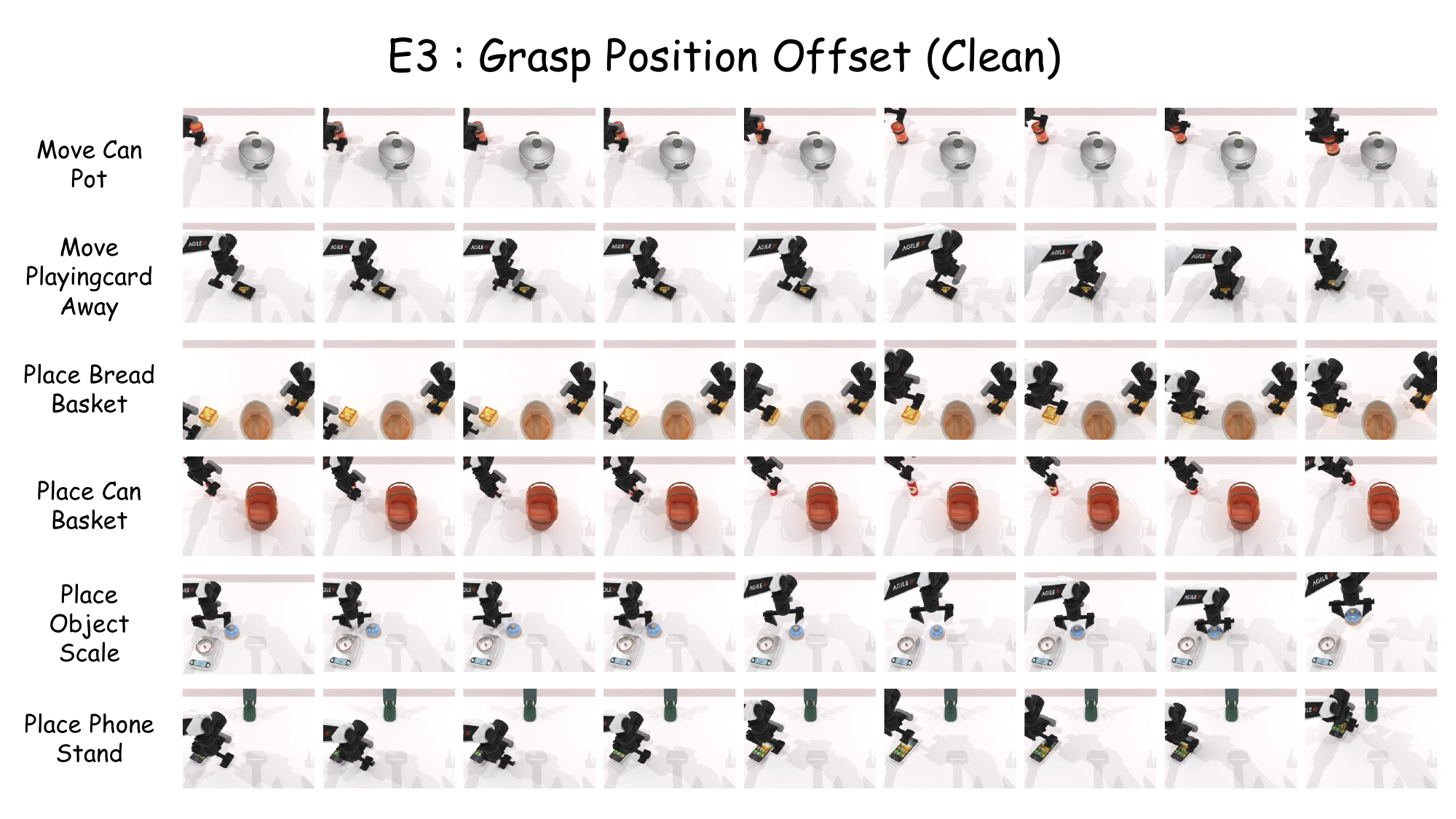}
    \label{fig:appendix_5}
    \caption{Grasp Position Offset (Clean)}
\end{figure*}

\begin{figure*}[htb]
    \centering
    \includegraphics[width=1.0\linewidth]{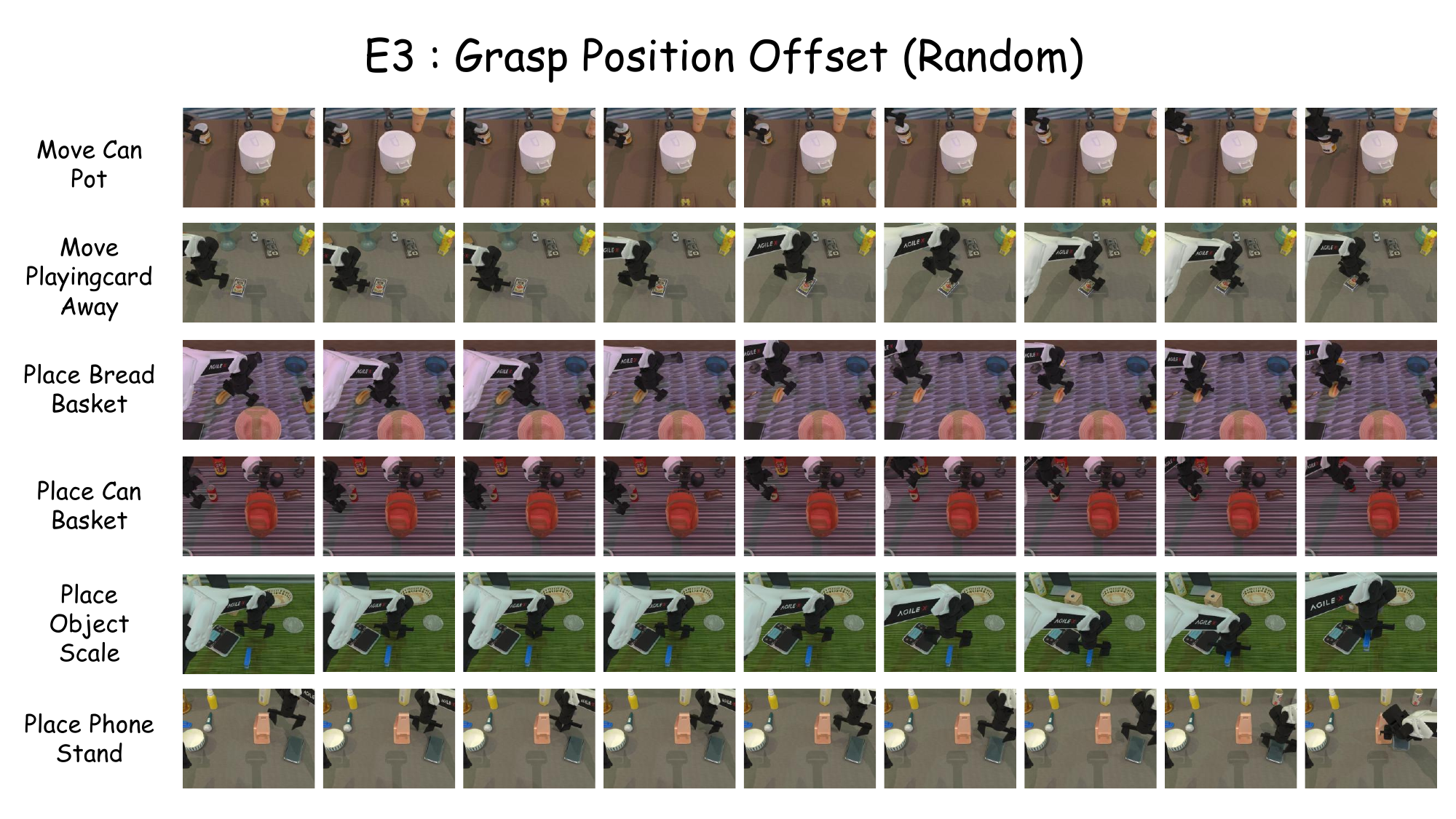}
    \label{fig:appendix_6}
    \caption{Grasp Position Offset (Random)}
\end{figure*}

\begin{figure*}[htb]
    \centering
    \includegraphics[width=1.0\linewidth]{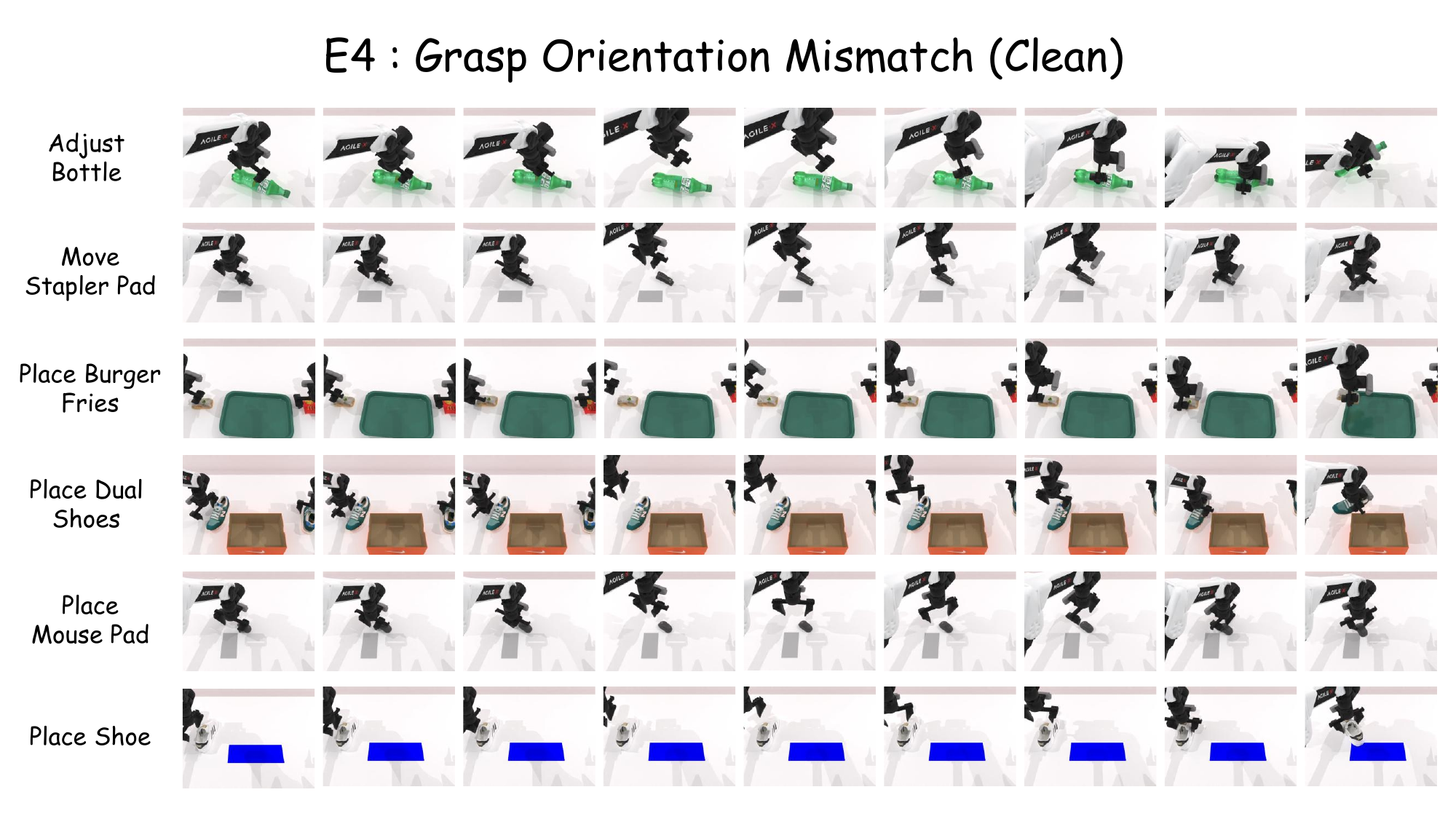}
    \label{fig:appendix_7}
    \caption{Grasp Orientation Mismatch (Clean)}
\end{figure*}

\begin{figure*}[htb]
    \centering
    \includegraphics[width=1.0\linewidth]{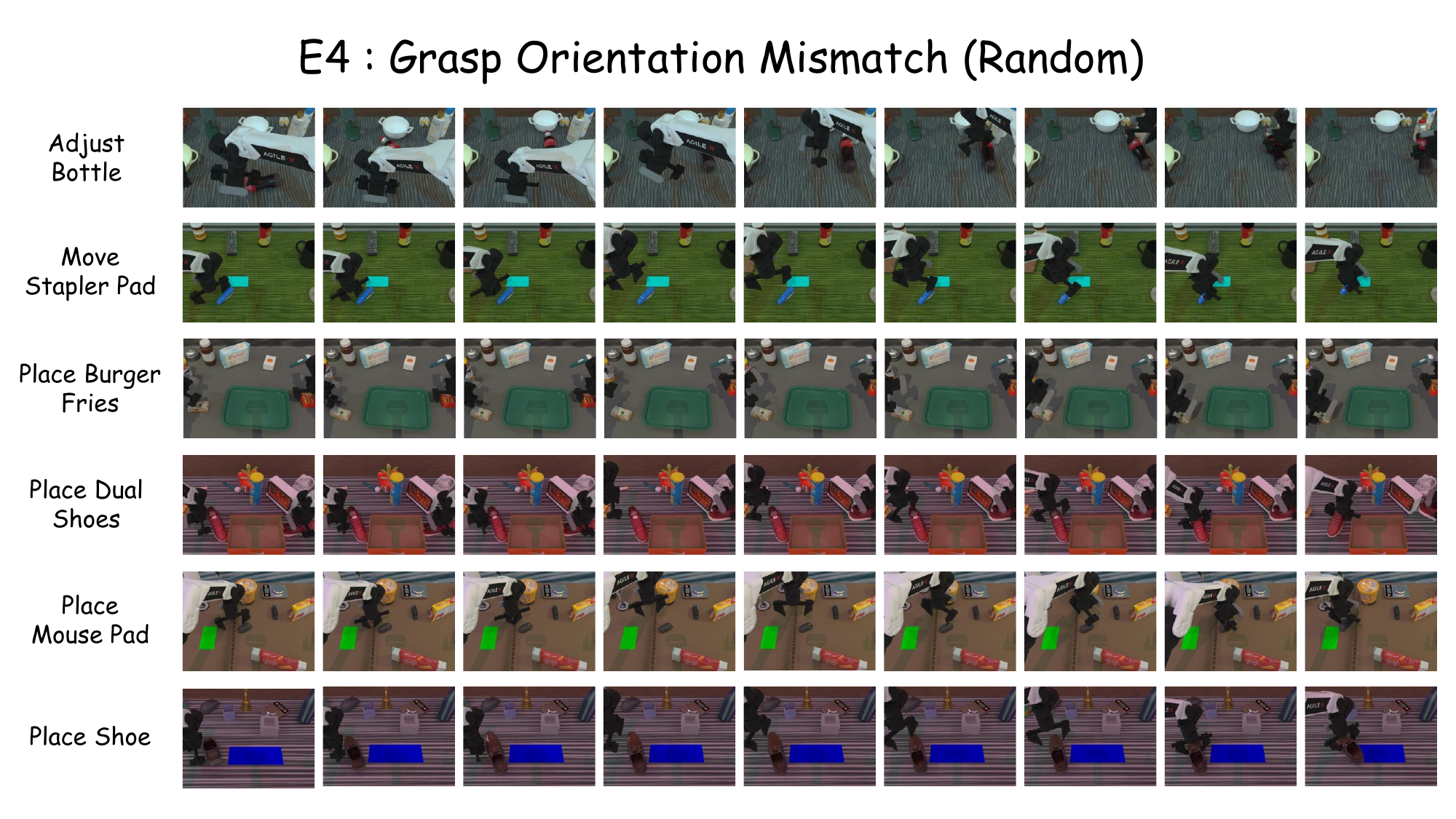}
    \label{fig:appendix_8}
    \caption{Grasp Orientation Mismatch (Random)}
\end{figure*}

\subsection{Qualitative Results: Real-World Evaluation}
\label{app:real_world_viz_sec}

\begin{figure*}[htbp]
    \centering
    \includegraphics[width=0.9\linewidth]{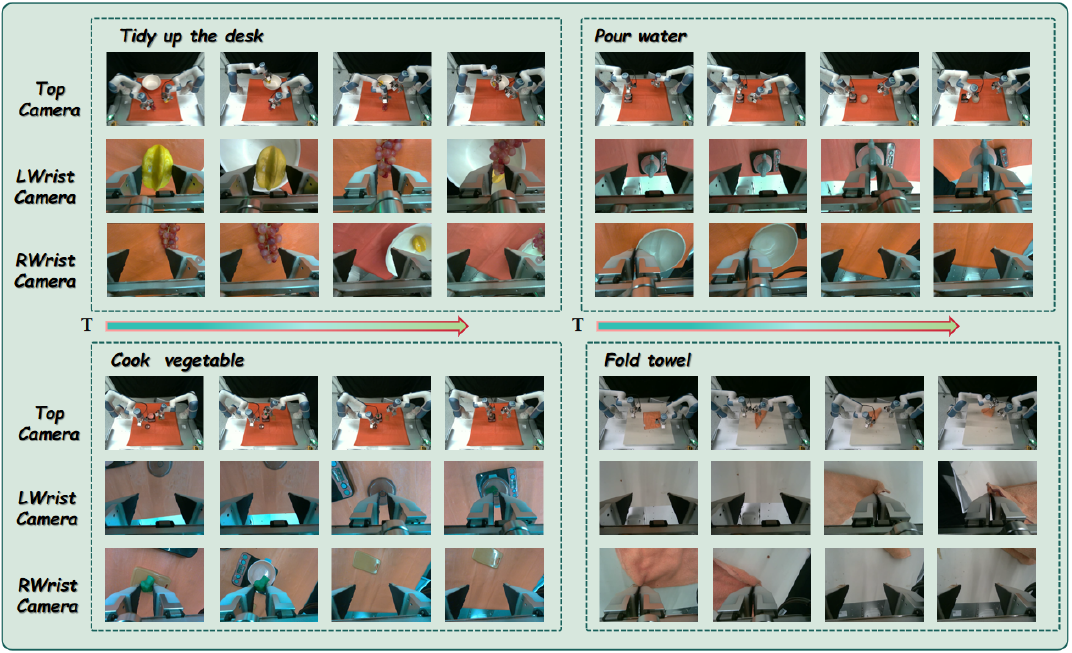}
    \caption{Real-World Evaluation Tasks. We evaluate our method on four contact-rich bimanual tasks. The figure showcases the environmental setup and key interaction phases for each task.}
    \label{fig:real_world_data}
\end{figure*}

In this section, we present qualitative visualizations of our real-world settings to complement the quantitative results in Section~\ref{sec:real_world_exp}. Figure~\ref{fig:real_world_data} illustrates the four challenging bimanual manipulation tasks used in our evaluation: \textit{Pour Water}, \textit{Cook Vegetable}, \textit{Tidy Desk}, and \textit{Fold Towel}. These tasks are selected to cover a wide range of manipulation primitives, including precision pouring, insertion, deformable object manipulation, and long-horizon coordination.

Comparing the execution traces under adversarial perturbation, RePO-VLA responds to forced deviations (e.g., external pushes or object slippage) by synthesizing valid recovery trajectories that resume task progress. In contrast, baseline methods often fail to react to these deviations, leading to task failures.

\end{document}